\documentclass[lettersize,journal]{IEEEtran}
\usepackage{amsmath,amsfonts}
\usepackage{algorithm}
\usepackage{algpseudocode}
\usepackage{array}
\usepackage[caption=false,font=normalsize,labelfont=sf,textfont=sf]{subfig}
\usepackage{textcomp}
\usepackage{stfloats}
\usepackage{url}
\usepackage{verbatim}
\usepackage{cite}
\usepackage{colortbl}
\usepackage[utf8]{inputenc} 
\usepackage[T1]{fontenc}    
\usepackage{hyperref}       
\usepackage{booktabs}       
\usepackage{threeparttable}
\usepackage{amsfonts}       
\usepackage{nicefrac}       
\usepackage{microtype}      
\usepackage{amsmath}
\usepackage{amssymb}
\usepackage{mathtools}
\usepackage{amsthm}
\usepackage{bbm}
\usepackage{tabularx}
\usepackage{arydshln}
\usepackage{adjustbox}
\usepackage{tabulary,multirow,overpic,xcolor}
\usepackage{algpseudocode}
\usepackage{graphicx}
\usepackage{amsthm}

\newtheorem{lemma}{Lemma}
\newtheorem{remark}{Remark}

\newcommand{\appref}[1]{Appendix~\hyperref[#1]{\ref*{#1}}}
\newcommand{\secref}[1]{Section~\hyperref[#1]{\ref*{#1}}}
\newcommand{\subsecref}[1]{Section~\hyperref[#1]{\ref*{#1}}}
\newcommand{\figref}[1]{Figure~\hyperref[#1]{\ref*{#1}}}
\newcommand{\tabref}[1]{Table~\hyperref[#1]{\ref*{#1}}}
\newcommand{\equref}[1]{Equation~\hyperref[#1]{\ref*{#1}}}

\newif\ifhighlight
\highlightfalse 

\long\def\hl#1{%
  \ifhighlight
    {\color{blue}#1}%
  \else
    #1%
  \fi
}

\begin{document}

\title{LM-SPT: LM-Aligned Semantic Distillation for Speech Tokenization}

\author{
Daejin Jo, Jeeyoung Yun, Byungseok Roh, and Sungwoong Kim
\thanks{Accepted for publication in IEEE/ACM Transactions on Audio, Speech, and Language Processing. 
© 2026 IEEE. Personal use of this material is permitted. 
Permission from IEEE must be obtained for all other uses, in any current or future media, 
including reprinting/republishing this material for advertising or promotional purposes, 
creating new collective works, for resale or redistribution to servers or lists, 
or reuse of any copyrighted component of this work in other works.}
\thanks{Daejin Jo, Jeeyoung Yun, and Sungwoong Kim are with the Department of Artificial Intelligence, Korea University, Seoul 02841, Republic of Korea (e-mail: twidddj@gmail.com; qwer010910@gmail.com; swkim01@korea.ac.kr).
Daejin Jo and Byungseok Roh are with the Multi-modal Model Training, Kakao Corp, Seongnam-si 13529, Republic of Korea. Corresponding author: Sungwoong Kim.}
}


\maketitle

\begin{abstract}
With the rapid progress of speech language models (SLMs), discrete speech tokens have emerged as a core interface between speech and text, enabling unified modeling across modalities. 
Recent speech tokenization approaches aim to isolate semantic information from low-level acoustics to better align with language models (LMs). 
In particular, previous methods use self-supervised learning (SSL) teachers such as HuBERT to extract semantic representations, which are then distilled into a semantic quantizer to suppress acoustic redundancy as well as capture content-related latent structures. 
However, these tokenizers often operate at relatively high frame rates, producing token sequences significantly longer than their textual counterparts and hindering seamless integration with pretrained LMs.
Although recent methods attempt to reduce the token rate by applying uniform average pooling to SSL features, this can over-smooth content-bearing regions and dilute the structural information, thereby potentially limiting the LM alignment.
To address this, we propose LM-SPT, an LM-aligned speech tokenization method based on semantic speech-resynthesis distillation.
Instead of directly matching teacher and student features via pooling, LM-SPT resynthesizes speech from semantic tokens only and minimizes the discrepancy between representations extracted from the original and resynthesized waveforms using a frozen, LM-aligned speech encoder.
This indirect supervision avoids rigid temporal alignment and encourages dedicated semantic units that are more semantically aligned with LMs under reduced frame rates.
Experimental results show that the proposed LM-SPT \hl{consistently outperforms} previous semantic-enhanced speech tokenizers when applied to SLMs for the tasks of automatic speech recognition and text-to-speech, even without compromising the speech reconstruction fidelity at the codec level.
The code release information and audio samples will be provided on the project page: \url{https://ku-agi.github.io/lmspt/}.
\end{abstract}


\section{Introduction}
\IEEEPARstart{L}{arge} language models (LLMs) have recently demonstrated remarkable capabilities in modeling and generating natural language. 
Motivated by this success, speech language models (SLMs) have emerged, aiming to model human speech in a similar generative fashion~\cite{gpt4o, minmo, moshi, Qwen3-Omni, voila2025, spiritlm, glm4voice, mimoaudio2025}. 
These models typically rely on discrete representations of speech, which can be broadly categorized into semantic tokens and acoustic tokens. 
Semantic tokens are derived from self-supervised models such as HuBERT~\cite{hubert} or Wav2vec2~\cite{schneider2019wav2vec}, capturing high-level linguistic content. 
Acoustic tokens, on the other hand, are produced by neural audio codecs like EnCodec~\cite{encodec} or DAC~\cite{DAC} and are optimized for preserving fine-grained audio details such as timbre and prosody.

While both speech representations have proven useful, neither is ideal in isolation for integrating with LLMs. 
Semantic tokens align well with text but often lose paralinguistic information necessary for high-fidelity speech synthesis. 
Acoustic tokens maintain audio quality but exhibit weak alignment with textual content, leading to errors such as missing or duplicated words in generated speech. 
Hierarchical approaches that combine both token types often suffer from high complexity, slower inference, and potential error accumulation.

To overcome these limitations, SpeechTokenizer~\cite{zhang2024speechtokenizer} proposes a unified tokenization framework that disentangles semantic and paralinguistic features within a single residual vector quantization (RVQ)~\cite{soundstream} architecture. 
The first RVQ layer learns semantic representations via explicit distillation from a semantic teacher (e.g., HuBERT), while subsequent layers model speaker-specific and prosodic variations.
Similarly, X-Codec~\cite{ye2025codec} concatenates SSL representations with waveform features before RVQ, and jointly optimizes acoustic and semantic reconstruction, rather than directly matching teacher and student features.
These designs enable a single tokenizer to support accurate text alignment and high-quality speech generation simultaneously, forming the basis of a unified SLM.

Despite these advances, they still operate at relatively high frame rates (e.g., 50Hz), producing token sequences that are substantially longer than text and therefore impeding efficient integration with pretrained LMs.
Consequently, recent studies have explored tokenizers operating at lower frame rates while maintaining high fidelity. 
Among them, Mimi~\cite{moshi} and DualCodec~\cite{dualcodec} are particularly promising, demonstrating strong reconstruction quality and encouraging results when integrated with LLM-based modeling.
Building on the semantic disentanglement principle of SpeechTokenizer, Mimi and DualCodec operate at a reduced frame rate of 12.5Hz, while their SSL teachers typically produce representations at a higher temporal resolution (e.g., 50Hz).
To align the temporal resolution of the teacher with the tokenizer for semantic distillation, a uniform average pooling is applied to downsample the teacher's representations to 12.5Hz.
However, this rigid frame-wise alignment neglects the fact that the importance of semantic information varies across time—compressing all segments equally can lead to disproportionate loss of critical content.
Furthermore, these methods depend on SSL teachers whose representations are shaped by phonetic regularities rather than higher-level semantic abstraction~\cite{wells2022phonetic,choi2024self}, which can constrain the type of linguistic structure transferred to discrete tokens~\cite{xu2024comparing}.

To address this, we propose LM-SPT\footnote{LM-SPT stands for \textbf{L}anguage-\textbf{M}odel\textbf{-}aligned \textbf{SP}eech \textbf{T}okenizer.}, an LM-aligned speech tokenization method that replaces SSL-based supervision with a pretrained \emph{LM-aligned speech encoder} (e.g., Whisper~\cite{whisper}) and introduces \emph{semantic speech-resynthesis distillation}.
Instead of directly matching pooled teacher and student features, LM-SPT resynthesizes speech solely from semantic tokens and minimizes the discrepancy between the teacher representations of the original and resynthesized waveforms.
To make this resynthesis-based distillation more effective, we further incorporate two key design choices: (i) we decouple semantic distillation from the main acoustic reconstruction pathway by introducing a lightweight \emph{auxiliary semantic decoder} dedicated to semantic speech-resynthesis, and (ii) we adopt a \emph{dual-encoder} architecture to mitigate interference between semantic and acoustic encoding objectives.
Together, these components encourage discrete units that are better aligned with LMs while maintaining high reconstruction fidelity, even under aggressive frame-rate reduction.

We empirically demonstrate that LM-SPT improves semantic alignment with LMs without degrading codec reconstruction fidelity. In particular, when integrated with LLMs, SLMs trained with LM-SPT tokens demonstrate strong downstream capability, achieving consistently improved performances on \textit{automatic speech recognition (ASR)} and \textit{text-to-speech (TTS)} compared to prior speech tokenizers. 
These results highlight that our LM-aligned, semantic speech-resynthesis distillation yields discrete speech units that remain high-fidelity at the codec level while being more suitable for SLMs.

Our main contributions can be summarized as follows:
\begin{itemize}
    \item We propose LM-SPT, a novel semantic-enhanced speech tokenization method, based on a \emph{semantic speech-resynthesis distillation} to learn discrete speech units that are more semantically aligned with LMs.

    \item We introduce a dual-encoder architecture with a separate decoder for speech resynthesis from semantic tokens, enabling effective semantic speech-resynthesis distillation.

    \item We benchmark speech tokenizers on downstream ASR and TTS tasks using SLMs, where LM-SPT consistently outperforms prior tokenizers, demonstrating that our learned tokens better support LLM-based speech understanding and generation.

    \item We evaluate the codec-level reconstruction on the Codec-SUPERB benchmark and show that LM-SPT matches upon the corresponding baselines, indicating that improved LM alignment does not come at the cost of reconstruction fidelity.

\end{itemize}

\begin{figure*}[]
    \centering
    \subfloat[Feature-level distillation]{%
        \includegraphics[width=0.47\linewidth]{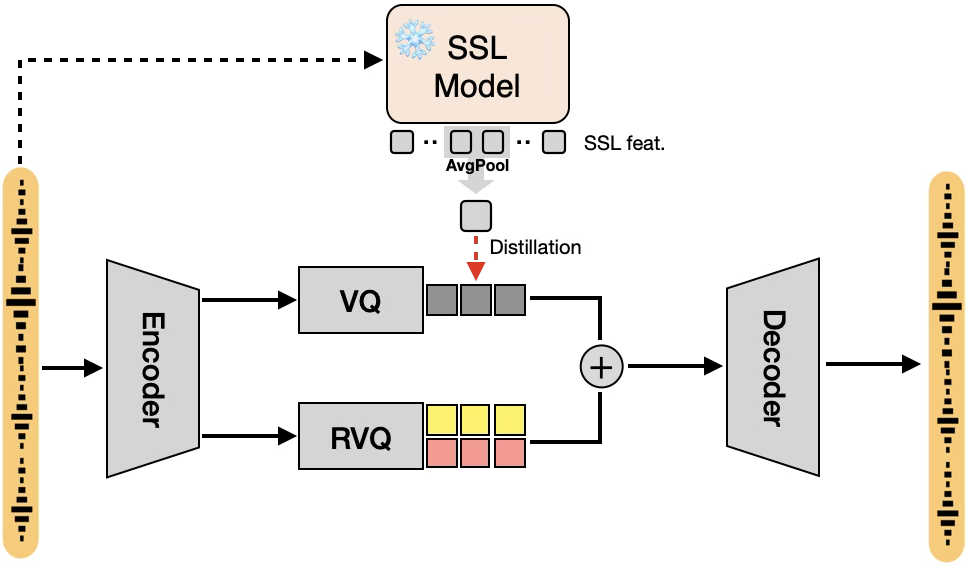}
        \label{fig:schematic_a}
    }
    \hspace{0.01\linewidth}
    {\color{gray!60}\rule{0.6pt}{0.22\textheight}} 
    \hspace{0.01\linewidth}
    \subfloat[Feature-level reconstruction distillation]{%
        \includegraphics[width=0.47\linewidth]{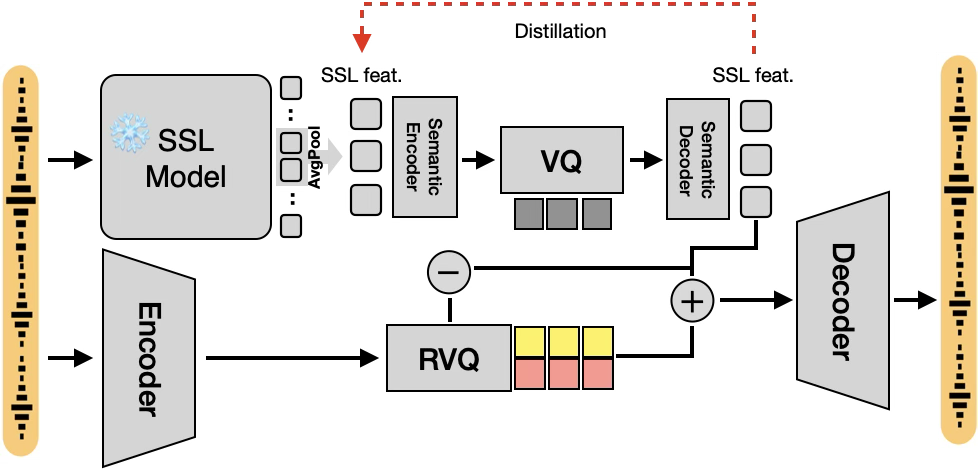}
        \label{fig:schematic_b}
    }

    \vspace{1em}

    \subfloat[Semantic speech-resynthesis distillation (LM-SPT)]{%
        \includegraphics[width=0.65\linewidth]{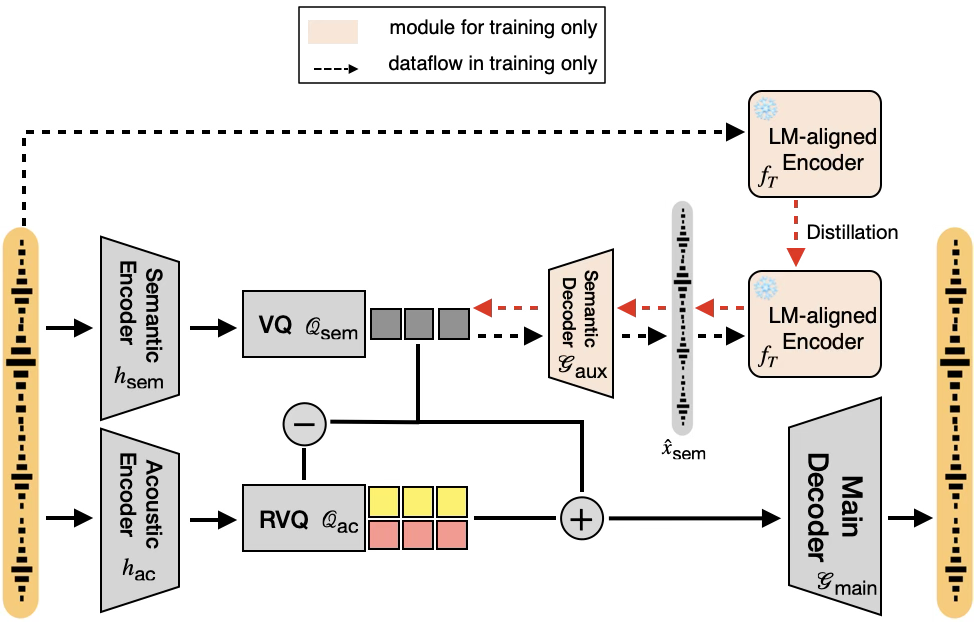}    
        \label{fig:schematic_c}
    }

    \vspace{1em}
    
    \caption{
    Comparison of semantic distillation approaches: (a) Feature-level distillation (SpeechTokenizer~\cite{zhang2024speechtokenizer}, Mimi~\cite{moshi}) matches the semantic pathway to SSL model features using frame-wise supervision (with temporal downsampling when operating at lower frame rates). (b) Feature-level reconstruction distillation (DualCodec~\cite{dualcodec}) reconstructs SSL model features through a semantic bottleneck, still relying on rigid frame-wise alignment. (c) Semantic speech-resynthesis distillation (LM-SPT) distills semantics by resynthesizing speech from semantic tokens only and minimizing the discrepancy between LM-aligned encoder features of the original and resynthesized waveforms. LM-SPT adopts a dual-encoder architecture, where separate encoders are trained for the semantic and acoustic quantizers.
    }
    \label{fig:schematic}
\end{figure*}

\section{Related Works} \label{sec:related}
\subsection{High-Fidelity Audio Codecs}
Advances in neural audio codecs have greatly enhanced speech tokenization and audio generation capabilities. EnCodec~\cite{encodec} and DAC~\cite{DAC} employ an encoder-decoder architecture utilizing RVQ and advanced adversarial training techniques.
WavTokenizer~\cite{wavtokenizer} further enhances codec efficiency by reducing the number of required tokens per second while maintaining subjective audio quality.
FACodec~\cite{facodec} employs Factorized Vector Quantization to explicitly separate distinct speech attributes, including content, prosody, acoustic details, and timbre, effectively supporting zero-shot speech synthesis tasks. However, these codecs primarily focus on acoustic reconstruction and compression efficiency, often overlooking explicit semantic modeling required by LMs.

\subsection{Semantic-Enhanced Speech Tokenizers}
CosyVoice~\cite{cosyvoice} introduces a supervised semantic speech tokenizer, which is trained to predict the transcript (textual content) directly from token sequences extracted from an input audio.
Similarly, GLM4-Voice tokenizer~\cite{glm4voice} finetunes Whisper~\cite{whisper} encoder with an additional pooling layer for frame rate reduction and a VQ layer for discretization.
Here, the supervision is focused primarily on semantic content rather than acoustic details.
As a result, decoding the encoded tokens into high-fidelity speech typically relies on powerful generative decoders such as diffusion- or flow-matching-based models, and often requires additional conditioning prompts to control speaker timbre during synthesis as discussed in~\cite{almtokenizer}.
In speech language modeling, this design may limit the expressiveness of the generated speech, since the language model does not explicitly generate acoustic tokens.
Moreover, such objectives tend to induce \emph{semantic--acoustic entangled} token representations, where token sequences may change substantially due to subtle acoustic factors even when the underlying semantic content is unchanged, which may make them harder for an LM to interpret and generalize~\cite{stabletoken}.

SpeechTokenizer~\cite{zhang2024speechtokenizer} leverages semantic distillation from a self-supervised learning (SSL) teacher (HuBERT~\cite{hubert}) at the high temporal resolution (50Hz), injecting distilled information into the first layer of an RVQ stack to improve semantic alignment while preserving acoustic detail in subsequent layers.
X-Codec~\cite{ye2025codec} also operates at 50Hz but adopts a different strategy: it concatenates SSL representations with waveform features before RVQ, and jointly optimizes acoustic and semantic reconstruction, rather than directly matching teacher and student features.
\hl{
In a complementary direction, RepCodec~\cite{repcodec} instead applies the codec encoder--decoder--VQ framework directly to SSL features themselves, learning discrete tokens that reconstruct continuous representations of a pretrained teacher.
}
Overall, such designs tend to induce \emph{semantic--enhanced} tokenizers and improve semantic alignment, which has recently been shown to offer advantages for integrating discrete speech tokens with language models~\cite{moshi,Qwen3-Omni,opusLM,mimoaudio2025,voila2025}.

\subsection{Low Frame Rate Speech Tokenizers}
Despite the advantages of the semantic-enhanced tokenizers, they still operate at relatively high frame rates (e.g., 50Hz), producing token sequences that are substantially longer than text and thus impeding efficient integration with pretrained LMs.
In addition, with the growing adoption of \textit{parallel decoding} in modern SLMs~\cite{moshi, Qwen3-Omni, voila2025, mimoaudio2025, opusLM, UALM}, where multiple codebooks are predicted in parallel at each time step, lowering the frame rate becomes one of the most important objectives for improving scalability, efficiency and LM alignment.
Consequently, recent studies have explored tokenizers operating at lower frame rates.
Notably, this line of work targets \emph{frame-rate} reduction—i.e., shortening the temporal token sequence length—rather than the conventional \emph{bit-rate} reduction emphasized in neural audio codec research, since frame rate directly determines the number of time steps in the token sequences.
Among recent low frame rate designs, \emph{Mimi}~\cite{moshi} and \emph{DualCodec}~\cite{dualcodec} are prominent semantic-enhanced tokenizers, both operating at 12.5Hz to compress the temporal sequence length.
Mimi leverages the feature-level supervision principle of SpeechTokenizer~\cite{zhang2024speechtokenizer}, using WavLM~\cite{wavlm} as the semantic teacher and adopting a split RVQ structure.
DualCodec builds a dual-stream codec architecture that separates semantic and acoustic information similar to X-Codec~\cite{ye2025codec}. 
It leverages a pretrained SSL model (w2v-BERT-2.0~\cite{w2v-bert}) as a dedicated semantic encoder whose representations are discretized by a semantic vector quantizer, alongside a separate acoustic encoder followed by an acoustic RVQ stack.
Training is performed with a semantic feature reconstruction objective for the semantic stream, while the semantic codebook embedding and the quantized acoustic codebook embedding are jointly used to reconstruct the input waveform, encouraging the codec to preserve both content-relevant and acoustic details.

In both cases, to bridge the mismatch between the low frame rate tokenizer and the high frame rate SSL teacher, the teacher representations are downsampled using uniform average pooling, enforcing a rigid frame-wise alignment for supervision.
This introduces two coupled issues under frame-rate reduction: (i) the SSL teachers used for distillation primarily encode phonetic/acoustic cues rather than higher-level semantics, and (ii) the pooled, one-to-one temporal alignment treats all time steps as equally important, which can over-smooth content-bearing regions and dilute the semantic structure in ways that may hinder effective integration with pretrained LMs for understanding and generation.
Additionally, when adopting a design similar to X-Codec or DualCodec, i.e., using a pretrained teacher model as a dedicated semantic encoder, the overall encoder footprint can become substantially larger. 
For instance, the SSL encoders used in X-Codec, HuBERT and WavLM, have about 95M parameters, while DualCodec relies on w2v-BERT-2.0, a 600M-parameter Transformer encoder. 
Moreover, beyond the pretrained backbone itself, these designs often require additional learnable modules (e.g., projection/adaptation layers) on top of the pretrained encoder, further increasing the total parameter count and runtime footprint.
In contrast, SpeechTokenizer and Mimi use comparatively lightweight codec encoders (67.7M and 37.8M parameters, respectively), making them substantially more amenable to efficient integration and deployment.

On the other hand, the proposed LM-SPT addresses these limitations by replacing SSL-based, pooled feature supervision with LM-aligned semantic speech-resynthesis distillation.
By supervising semantic tokens through resynthesized waveforms and a frozen LM-aligned speech encoder, LM-SPT avoids rigid frame-wise alignment and remains effective at low frame rates.
Moreover, LM-SPT does not require a large pretrained SSL model as a dedicated semantic encoder, enabling a lightweight tokenizer design that preserves reconstruction fidelity while producing discrete units that are more suitable for pretrained LMs in both understanding and generation.

\hl{
Outside this pretrained teacher-distillation-based family, Single-Codec~\cite{singlecodec} also targets low frame rate ($\sim$23Hz) tokenization for LLM-based TTS, but achieves this via mel-spectrogram-domain disentanglement of time-invariant information into a continuous reference embedding rather than via semantic distillation.
}

\section{Method}
\subsection{Motivation}
Prior works such as SpeechTokenizer~\cite{zhang2024speechtokenizer} and Mimi~\cite{moshi} learn discrete semantic representations by distilling continuous features from self-supervised learning (SSL) models (e.g., HuBERT~\cite{hubert} or WavLM~\cite{wavlm}) into a semantic quantizer. 
Specifically, a pretrained SSL model serves as a teacher, and its frame-level hidden representations are used as supervision targets for training a student encoder, as depicted in \figref{fig:schematic_a}.
DualCodec~\cite{dualcodec} introduces an explicit reconstruction step in the teacher feature space, as illustrated in \figref{fig:schematic_b}.
Concretely, the SSL teacher (w2v-BERT-2.0~\cite{w2v-bert}) provides frame-level representations, and the semantic quantizer is trained to preserve them by reconstructing the teacher features through an auxiliary semantic feature reconstruction path.

While effective in guiding the model toward semantically rich tokens, these approaches have the following limitations:
First, the SSL teacher’s representation may not align well with LM semantics, as it is trained to capture phonetic regularities rather than semantic abstraction~\cite{choi2024self,wells2022phonetic}.
Second, feature-level supervision implicitly requires the teacher and student to be defined on a comparable time grid. 
In practice, SSL teachers typically operate at a fixed high frame rate and are non-trivial to re-train or re-configure at lower rates; thus, previous low-rate tokenizers rely on heuristic temporal alignment to match resolutions, which can weaken supervision under frame-rate reduction.
Mimi and DualCodec follow this approach by distilling 12.5Hz representations from their 50Hz teacher, using average pooling to align the temporal resolution.
However, this temporal smoothing remains a rigid approximation and can distort or dilute semantic specificity required for effective alignment with language models.

\subsection{Semantic Speech-Resynthesis Distillation}
To better align semantic supervision with LM-level representations, we propose to use a pretrained LM-aligned speech encoder (e.g., Whisper~\cite{whisper}) as the semantic teacher. 
We therefore refer to our approach as \textit{LM-aligned semantic distillation}, as LM-aligned  representations serve as an effective proxy for language model semantics in practice~\cite{choi2024self, xu2024comparing}.

To overcome the challenge of temporal misalignment under low frame rates, we propose a semantic speech-resynthesis distillation strategy that bypasses direct feature-level supervision and instead employs an audio speech-resynthesis based alignment across different frame rates between teacher and student. 
In specific, as depicted in \figref{fig:schematic_c}, we reconstruct speech solely from semantic tokens and supervise the tokenizer by minimizing the distance between LM-aligned speech encoder features of the original and reconstructed speech waveforms. 
Formally, our semantic distillation loss is defined as:
\begin{align}
\label{eq:distill}
\begin{split}
    \mathcal{L}_\text{distill}^\text{\hl{resyn}} &= \lVert f_T(x) - f_T(\hat{x}_\text{sem}) \rVert^2,
\end{split}
\end{align}
where $f_T$ denotes the frozen, off-the-shelf LM-aligned speech encoder, which maps waveforms $x$ to semantic representations. 
Using an auxiliary decoder $\mathcal{G}_\text{aux}$, the reconstructed waveform $\hat{x}_\text{sem} = \mathcal{G}_\text{aux}(z_\text{sem})$ is generated from semantic tokens $z_\text{sem} = \mathcal{Q}_\text{sem}(h_\text{sem}(x))$, where $\mathcal{Q}_\text{sem}$ is the semantic vector quantization module applied to the semantic encoder output $h_\text{sem}(x)$.
\hl{
We defer a theoretical justification of this resynthesis-based distillation to~\secref{sec:theory}.
}

\subsection{Decoupling Semantic Distillation from the Main Codec Decoding} \label{method:decoup}
We emphasize that an \emph{auxiliary semantic decoder} $\mathcal{G}_\text{aux}$ is used for the semantic speech-resynthesis, without relying on acoustic tokens or any additional supervision. 
Namely, we do not share this decoder with the main reconstruction decoder $\mathcal{G}_{\text{main}}$. 
This design choice is based on the following considerations:
First, using a shared decoder for both audio reconstruction and semantic distillation may lead to interference between tasks, as the decoder is jointly optimized for potentially conflicting objectives. 
Second, one might attempt to mitigate this interference by freezing the decoder during distillation, allowing it to focus solely on reconstruction. 
While this can improve fidelity by preventing gradient conflict, we find that it also weakens the effect of distillation, as the decoder no longer adapts to match semantic supervision, which is confirmed through empirical analysis (see \subsecref{sec:abl_decoder}).
Moreover, we deliberately design this decoder to have limited capacity, functioning as an information bottleneck.
This constraint forces the semantic encoder to produce more expressive and compact representations. 
In \subsecref{sec:abl_decoder}, we demonstrate that using a lightweight decoder indeed leads to stronger semantic representations under LM-alignment evaluation.

\begin{figure*}[]
    \centering
    \subfloat[ASR]{%
        \includegraphics[width=0.28\linewidth]{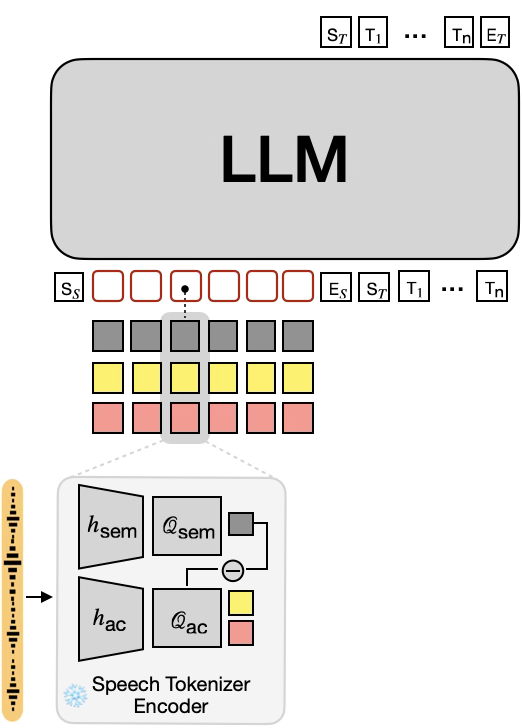}
        \label{fig:slm_asr}
    }
    \hspace{0.01\linewidth}
    {\color{gray!60}\rule{0.6pt}{0.22\textheight}} 
    \hspace{0.01\linewidth}
    \subfloat[TTS]{%
        \includegraphics[width=0.62\linewidth]{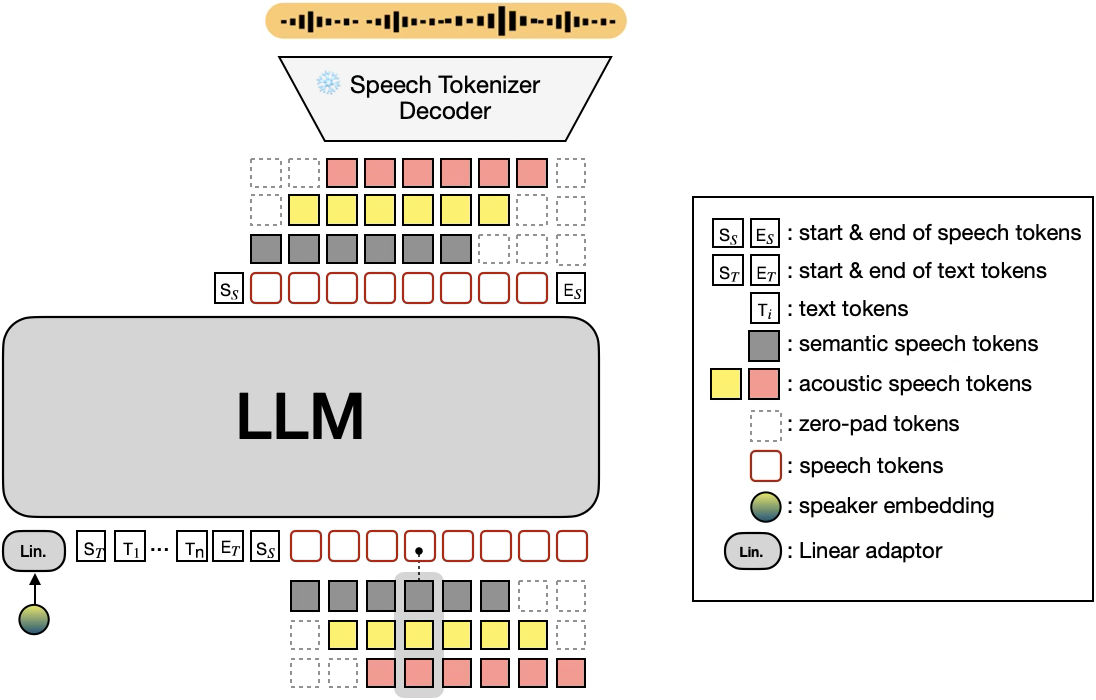}
        \label{fig:slm_tts}
    }
    \vspace{0.5em}
    \caption{
    Speech Language Models (SLMs) architecture for (a) Automatic speech recognition (ASR) and (b) Text-to-Speech (TTS). Delay pattern ($\text{delay}=1$) is applied to speech tokens in TTS task. Raw audio waveform is discretized to speech tokens (gray for semantic and colored for acoustic) by the speech tokenizer encoder and quantizer. Each discrete token is then mapped to an input embedding via a lookup table initialized randomly and learned during training, and fed into the LLM in the same way as text token embeddings.  For tokenizers with multiple codebooks, the semantic-token embedding and acoustic-token embeddings at the same time step are summed to form the final speech-token input embedding.
    }
    \label{fig:SLM}
\end{figure*}

\subsection{Split RVQ with Dual Encoder} \label{method:dual_enc}
Following recent semantic-enhanced speech tokenizers~\cite{zhang2024speechtokenizer,moshi,dualcodec}, we adopt a similar split quantization architecture in LM-SPT, consisting of a dedicated semantic vector quantizer $\mathcal{Q}_\text{sem}$ and a residual vector quantizer stack $\mathcal{Q}_\text{ac}$ for acoustic encoding.
Moreover, inspired by prior works on disentangled representation learning~\cite{qu2024tokenflow,ye2025codec}, we adopt a dual encoder architecture, where the semantic and acoustic encoders are separately optimized to capture high-level semantic information and low-level acoustic details, respectively.  
In contrast to SpeechTokenizer~\cite{zhang2024speechtokenizer} and Mimi~\cite{moshi}, which either use a shared encoder or partially disentangled semantic and acoustic signals via quantization, our model explicitly separates the encoding process at the architectural level.
Acoustic vector quantization is applied to the acoustic encoder output, defined as $z_\text{ac} = \mathcal{Q}_\text{ac}(h_\text{ac}(x) - z_\text{sem})$, where $h_\text{ac}(x)$ is the output of the acoustic encoder.
We use the same architecture for both encoders.

Note that X-Codec~\cite{ye2025codec} and DualCodec~\cite{dualcodec} also introduce dual semantic/acoustic streams. 
However, their semantic stream is explicitly conditioned on features extracted from a large pretrained SSL encoder (e.g., HuBERT/WavLM or w2v-BERT-2.0), which must be kept as an external dependency for token extraction. 
In contrast, LM-SPT does not require a pretrained semantic encoder as a component of the tokenizer: both $h_{\text{sem}}$ and $h_{\text{ac}}$ are learned from scratch, while a frozen LM-aligned speech encoder is used only during training to provide a distillation objective.
In~\subsecref{sec:abl_encoder}, we empirically show that this dual encoder improves both reconstruction fidelity and semantic alignment, particularly when compared to models using a shared encoder.

\subsection{Training Objective}
In this study, we implement LM-SPT on top of DAC~\cite{DAC} and build LM-SPT upon the DualCodec training framework~\cite{dualcodec}.
We retain the original set of codec training losses from the framework, including the spectrogram reconstruction loss $\mathcal{L}_{\text{spec}}$, adversarial loss $\mathcal{L}_{\text{adv}}$, quantization loss $\mathcal{L}_{\text{vq}}$, commitment loss $\mathcal{L}_{\text{comm}}$, and feature matching loss $\mathcal{L}_{\text{feat}}$.
We train LM-SPT by minimizing a weighted sum of its constituent loss components:
\begin{align*}
    \mathcal{L}_G 
    &= \lambda_{\text{spec}}\,\mathcal{L}_{\text{spec}}
      + \lambda_{\text{adv}}\,\mathcal{L}_{\text{adv}}
      + \lambda_{\text{feat}}\,\mathcal{L}_{\text{feat}} 
      + \lambda_{\text{vq}}\,\mathcal{L}_{\text{vq}} \\
      & \qquad + \lambda_{\text{comm}}\,\mathcal{L}_{\text{comm}}
      + \lambda_{\text{distill}}\,\mathcal{L}_{\text{distill}}^\text{\hl{resyn}}.
\end{align*}
We set $\lambda_{\text{spec}}=1$, $\lambda_{\text{adv}}=1$, $\lambda_{\text{feat}}=1$, $\lambda_{\text{vq}}=1$, $\lambda_{\text{comm}}=0.25$, and $\lambda_{\text{distill}}=500$ throughout the experiments.

\bgroup
\def\arraystretch{1.2}%
\begin{table*}[t]
\resizebox{\textwidth}{!}{
    \begin{threeparttable}
    \centering
    \caption{\label{tab:main_recon} Codec Performance Comparison. Signal-level and Application-level scores are obtained from Codec-SUPERB~\cite{codec-superb}. UNMI denotes utterance-normalized mutual information (see details in~\subsecref{sec:exp_unmi}).}
    \small

        \begin{tabular}{lllccccccccc}
            \toprule
            & & & \multicolumn{3}{c}{\textit{Signal-level}} & \multicolumn{5}{c}{\textit{Application-level}} & {\textit{Semantic-level}} \\
            \cmidrule(lr){4-6}\cmidrule(lr){7-11}\cmidrule(lr){12-12}
            Frame rate & Model & \#Params & SDR $\uparrow$ & Mel loss $\downarrow$ & PESQ $\uparrow$ & ACC (ER) $\uparrow$ & EER (ASV) $\downarrow$ & \multicolumn{2}{c}{WER (ASR)} $\downarrow$ & ACC (EC) $\uparrow$ & UNMI $\uparrow$ \\ 
            \cmidrule(lr){9-10}
            & & & & & & & &
            \multicolumn{1}{c}{\scriptsize \textit{Whisper}} &
            \multicolumn{1}{c}{\scriptsize \hl{\textit{Wav2Vec2-En}}} &
            & \\
            \midrule
            \multicolumn{11}{l}{\textbf{\textit{High-fidelity audio codecs}}}\\ 
            75Hz & EnCodec~\cite{encodec} & 15M & 7.50 & 1.03 & 2.48 & 70.28 & 2.44 & 3.60 & \hl{7.12} & 87.75 & -- \\
            75Hz & DAC~\cite{DAC} & 75M & 3.62 & 0.68 & 2.98 & 71.88 & 2.33 & 3.42 & \hl{6.63} & 84.45 & -- \\
            40Hz & WavTokenizer~\cite{wavtokenizer}  & 81M & -9.96	 & 1.11	 & 1.39 & 60.07 & 17.99 & 20.35 & \hl{24.73} & 23.30 & -- \\
            \midrule
            \multicolumn{11}{l}{\textbf{\textit{Semantic-enhanced speech tokenizers}}}\\
            50Hz & SpeechTokenizer~\cite{zhang2024speechtokenizer}  & 103M & 1.52 & 0.84 & 1.98 & \underline{74.17} & 3.19 & 4.42 & \hl{6.26} & 50.90 & 0.7403 $\pm 0.019$ \\
            25Hz & CosyVoice2\dag~\cite{cosyvoice2} & 257M & -15.22 & 1.87 &1.17 & 58.61 & 21.04 & 7.44 & \hl{10.20} & 31.75 & 0.7507 $\pm 0.018$ \\
            12.5Hz & Mimi~\cite{moshi}  & 79M & 2.88 & 1.17 & 1.87 & 69.24 & 4.82 &	6.24 & \hl{10.14} & \textbf{64.45} & 0.7270 $\pm 0.026$ \\
            12.5Hz & DualCodec~\cite{dualcodec}  & 665M & \underline{4.48} & \underline{0.78} & \textbf{2.37} & \textbf{74.65} & \underline{2.81} & \underline{3.71} & \hl{\textbf{5.44}} & 55.40 & \underline{0.7757} $\pm 0.007$\\
            12.5Hz & LM-SPT & 74M & \textbf{5.54} & \textbf{0.75} & \underline{2.21} & 73.54 & \textbf{2.67} & \textbf{3.65} & \hl{\underline{5.47}} & \underline{56.55} & \textbf{0.7925} $\pm 0.004$\\
            \bottomrule
        \end{tabular}

    \begin{tablenotes}
        \item[\dag] CosyVoice2 is a supervised semantic-token-based synthesis model rather than a general-purpose neural audio codec optimized for audio reconstruction, its codec-benchmark scores should be interpreted accordingly.
        \item[1] \textit{The \textbf{bold} text and \underline{underscored} text represent the best and second-best results.}
    \end{tablenotes}
    \end{threeparttable}
}
\end{table*}
\egroup

\section{Experiments} \label{sec:experiments}
\subsection{Speech Tokenizer Setup}
We adopt the DAC~\cite{DAC} architecture, consisting of a codec encoder, an RVQ, and a codec decoder.
Following the 12.5Hz version of DualCodec setup, we use the same acoustic encoder and decoder configurations: the encoder is composed of 5 CNN blocks with strides $(4,5,6,8,2)$, and the decoder mirrors the encoder with upsampling convolutions.
For LM-SPT, we employ a dual-encoder design where the semantic encoder $h_{\text{sem}}$ and the acoustic encoder $h_{\text{ac}}$ have the same architecture; all parameters of LM-SPT are randomly initialized and trained from scratch.
We use a semantic quantizer with a codebook size of $16{,}384$ and an acoustic RVQ stack with 7 codebooks of size $4{,}096$ each. 
The codebook embedding dimension is set to 8 for both the semantic and acoustic quantizers, and we apply acoustic quantizer dropout~\cite{DAC} with probability 0.5.
For semantic speech-resynthesis, the auxiliary decoder $\mathcal{G}_{\text{aux}}(z_{\text{sem}})$ adopts the same overall architecture as the main decoder $\mathcal{G}_{\text{main}}$ but is configured as a lightweight upsampling stack with strides $(8,8,5,4)$, producing 16~kHz waveforms (i.e., $12.5 \times 8 \times 8 \times 5 \times 4$). 
We set its number of channels to 128, resulting in an approximately 1.2M-parameter auxiliary decoder, which is substantially smaller than the main codec decoder (40.5M parameters).
We optimize all models with AdamW using $\beta_1=0.8$ and $\beta_2=0.9$, following DualCodec. 
The learning rate is scheduled with cosine decay from $3\times10^{-4}$ to a minimum of $1\times10^{-5}$.

We use the Emilia dataset~\cite{emilia}, which contains approximately 100K-hour multilingual speech at a 24kHz sampling rate, and 64 NVIDIA A100 GPUs for training.
LM-SPT is trained for 3 epochs using randomly cropped 6-second audio segments at 24~kHz sampling rate. 
We use \textit{Whisper-large-v3}~\cite{whisper} as the pretrained LM-aligned speech encoder for distillation.

\subsection{Speech Language Model Setup}
To evaluate how well each speech tokenizer integrates with pretrained language models (PLM), we train PLM on two representative downstream tasks of \textit{speech language model (SLM)}: \textit{Automatic Speech Recognition (ASR)} as an understanding task, and \textit{Text-to-Speech (TTS)} as a generation task.
For TTS, we focus on a zero-shot speaker-conditioned setting to assess generalization to unseen speakers.

For ASR, we construct a bilingual (Korean and English) training set of approximately 173K hours by randomly sampling from multiple datasets: for English, we sample about 87K hours from GigaSpeech~\cite{GigaSpeech}, MLS~\cite{MLS}, and PeopleSpeech~\cite{PeopleSpeech}; for Korean, we sample about 86K hours from AIHub\footnote{\url{https://aihub.or.kr/}} and in-house data.
We use \textit{Qwen2.5-3B-Instruct}~\cite{qwen25} as the backbone LM.
The SLM architecture for the ASR task is depicted in \figref{fig:slm_asr}.

For TTS, we sample 12K hours of English speech from LibriHeavy~\cite{libriheavy} and 11K hours of Korean speech from AIHub, and use \textit{Qwen2.5-0.5B-Instruct}~\cite{qwen25} as the backbone LM as in~\cite{cosyvoice2}.
To enable zero-shot speaker conditioning, we extract a speaker embedding using a speaker verification model (\textit{ERes2NetV2}~\cite{3d-speaker}), project it to the LM input dimension, and prepend it as a prefix token to the transcript. The model is then trained to generate speech tokens conditioned on both the transcript and the speaker embedding.
Specifically, the speaker embedding is projected to the LM input dimension via a single learnable linear layer and then inserted as a prefix token to the textual prompt.
We utilize the parallel decoding with one-frame delay following~\cite{copet2023simple,mimoaudio2025,opusLM,UALM}.
The padding tokens introduced by the one-frame delay are zeroed out when fed into the LM.
For tokenizers that use multiple codebooks (e.g., Mimi, DualCodec, and LM-SPT), we use a multi-head output with one semantic token head and seven acoustic token heads, corresponding to the semantic codebook and the seven acoustic codebooks, respectively.
The SLM architecture for the TTS task is depicted in \figref{fig:slm_tts}.
We train both ASR and TTS SLMs using sequence packing~\cite{seqpack} with a sequence length of 16K on 32 NVIDIA A100 GPUs, running 50K steps for ASR and 15K steps for TTS.

\subsection{Speech Tokenizer Performance} \label{sec_exp:codec}
We report speech tokenizer performances using Codec-SUPERB~\cite{codec-superb}, which evaluates neural audio codecs along two complementary axes: \textit{signal-level} fidelity and \textit{application-level} utility.
While Codec-SUPERB covers multiple audio domains beyond speech, including music and general sound, our focus is on speech tokenization and speech codecs; therefore, we report results only on the speech subset of the benchmark.
At the signal level, the benchmark measures how faithfully the codec reconstructs the waveform using objective metrics such as temporal distortion with \textit{Signal-to-distortion ratio (SDR)} and perceptual quality with \textit{Mel loss}~\cite{DAC} and \textit{PESQ}~\cite{pesq}, summarizing overall reconstruction fidelity independent of any downstream model.
At the application level, Codec-SUPERB evaluates how well the reconstructed audio preserves information required by downstream tasks on the codec outputs.
This evaluation encompasses content (\textit{word error rate (WER)} for \textit{automatic speech recognition (ASR)}), speaker timbre (\textit{equal error rate (EER)} for \textit{automatic speaker verification (ASV)}), emotion
(accuracy (ACC) for \textit{speech emotion recognition (ER)}), and general audio characteristics (accuracy for \textit{audio event classification (EC)}).
\hl{
    For ASR evaluation, Codec-SUPERB uses \textit{Whisper-large-v3}~\cite{whisper} by default.
    In addition, we also report ASR results measured with a Wav2Vec2~\cite{wav2vec2}-based ASR model\footnote{\label{fn:w2v2en}\url{https://huggingface.co/facebook/wav2vec2-large-960h-lv60-self}}, which is denoted as \textit{Wav2vec2-En} in our tables.
}
The reconstruction results are summarized in \emph{Signal-level} and \emph{Application-level} columns of \tabref{tab:main_recon} and show that LM-SPT achieves competitive performances compared to the strong baseline DualCodec~\cite{dualcodec}, despite having substantially fewer parameters.

\bgroup
\def\arraystretch{1.2}%
\begin{table*}[t]
\resizebox{\textwidth}{!}{
    \begin{threeparttable}
    \centering
    \caption{\label{tab:main_slm} Downstream Performance of SLMs.}
    \small

    \begin{tabular}{l c cc cc ccccc ccccc}
    \toprule
    \multirow{4}{*}{Model} &
    \multirow{4}{*}{\begin{tabular}[c]{@{}c@{}}Frame rate (Hz)\\ / Codebooks\end{tabular}} &
    \multicolumn{4}{c}{\emph{Automatic Speech Recognition}} &
    \multicolumn{10}{c}{\emph{Zero-shot Text-to-Speech}} \\
    \cmidrule(lr){3-6}\cmidrule(lr){7-16}
    & &
    \multicolumn{2}{c}{LibriSpeech~\cite{librispeech}} &
    \multicolumn{2}{c}{KSponSpeech~\cite{KsponSpeech}} &
    \multicolumn{5}{c}{LibriSpeech (\textit{test-clean})} &
    \multicolumn{5}{c}{KSponSpeech (\textit{test-clean})} \\
    \cmidrule(lr){3-4}\cmidrule(lr){5-6}\cmidrule(lr){7-11}\cmidrule(lr){12-16}
    & &
    WER $\downarrow$ & WER $\downarrow$ & CER $\downarrow$ & CER $\downarrow$ &
    \multicolumn{2}{c}{WER $\downarrow$} & \multirow{2}{*}{UTMOS $\uparrow$} & \multirow{2}{*}{DNSMOS $\uparrow$} & \multirow{2}{*}{SIM $\uparrow$} &
    \multicolumn{2}{c}{CER $\downarrow$} & \multirow{2}{*}{UTMOS $\uparrow$} & \multirow{2}{*}{DNSMOS $\uparrow$} & \multirow{2}{*}{SIM $\uparrow$} \\
    \cmidrule(lr){7-8}\cmidrule(lr){12-13}
    & &
    \textit{test-clean} & \textit{test-other} & \textit{test-clean} & \textit{test-other} &
    \multicolumn{1}{c}{\scriptsize \textit{Whisper}} &
    \multicolumn{1}{c}{\scriptsize \hl{\textit{Wav2vec2-En}}} &
    & & &
    \multicolumn{1}{c}{\scriptsize \textit{Whisper}} &
    \multicolumn{1}{c}{\scriptsize \hl{\textit{Wav2vec2-Ko}}} &
    & & \\
    \midrule

    CosyVoice2~\cite{cosyvoice2} & 25.0 / 1
    & 35.27 & 45.61 & 39.82 & 42.18
    & \underline{3.08} & \hl{\underline{2.14}} & 3.78 & \textbf{3.41} & 0.52
    & \underline{9.98} & \hl{\underline{27.42}} & 2.77 & \textbf{3.35} & 0.42 \\
    Mimi~\cite{moshi} & 12.5 / 8
    & \underline{3.82} & 13.65 & 15.48 & 15.2
    & 3.29 & \hl{2.86} & 3.59 & 3.35 & \textbf{0.78}
    & 11.03 & \hl{36.30} & 2.20 & 3.15 & 0.66 \\
    DualCodec~\cite{dualcodec} & 12.5 / 8
    & 3.95 & \underline{12.13} & \underline{12.92} & \underline{12.93}
    & 4.62 & \hl{5.85} & \underline{4.14} & 3.34 & 0.76
    & 10.37 & \hl{33.12} & 2.98 & \underline{3.19} & \underline{0.74} \\
    LM-SPT & 12.5 / 8
    & \textbf{3.39} & \textbf{11.21} & \textbf{11.91} & \textbf{11.37}
    & \textbf{1.48} & \hl{\textbf{1.69}} & \textbf{4.37} & \textbf{3.41} & \underline{0.77}
    & \textbf{6.64} & \hl{\textbf{26.98}} & \textbf{3.08} & 3.18 & \textbf{0.75} \\

    \bottomrule
    \end{tabular}

    \begin{tablenotes}
        \item[1] \textit{The \textbf{bold} text and \underline{underscored} text represent the best and second-best results.}
    \end{tablenotes}
    \end{threeparttable}
}
\end{table*}
\egroup

\subsection{Utterance-level Consistency} \label{sec:exp_unmi}
To measure semantic alignment among codecs, we draw inspiration from~\cite{hubert}, which introduced \textit{Phoneme-Normalized Mutual Information (PNMI)} to quantify the consistency between phoneme units and learned discrete tokens. 
PNMI measures the mutual information (MI) between phonemes and token assignments, normalized by the entropy of the phoneme distribution.
However, \hl{as observed in~\cite{repcodec}}, phoneme-level alignment is not always a sufficient proxy for \emph{utterance-level linguistic content} in semantic-acoustic disentangled codecs: high MI with phoneme labels mainly reflects local phonetic consistency, while content consistency is often expressed over longer contexts.
Motivated by this, we shift the alignment target from phoneme units to \emph{sequence-level} units, and use the resulting score as a measure of how consistently discrete semantic tokens preserve \emph{utterance-level linguistic content}.

In this study, we devise \textit{Utterance-Normalized Mutual Information (UNMI)}, which simply replaces phoneme units with full utterance-level sequences as the alignment target.
Unlike phoneme sets, sequence-level units span a vast and diverse space.
To make the computation tractable, we utilize the VCTK dataset~\cite{vctk}, which contains approximately 44K utterances from 110 speakers reading a shared set of 400 sentences. 
This design enables us to restrict the semantic variation by leveraging repeated textual content across speakers.

For our UNMI evaluation, we randomly sample 18K utterances (approximately 40\% of the corpus) from VCTK.
Since each utterance yields a variable-length token sequence, we apply \textit{Locality Sensitive Hashing} to map each sequence into a discrete unit, which we treat as a proxy for sequence identity.
We set the hash size to $n=\lfloor\log_2(|\mathcal{V}_{\text{sem}}|)\rfloor $ bits, where $|\mathcal{V}_{\text{sem}}|$ denotes the vocabulary size of the semantic codebook.
We then compute mutual information between the hashed sequence identities and the associated ground-truth transcript identities, and report the mean and standard deviation over 10 runs with different random seeds in the UNMI column of \tabref{tab:main_recon}.
We observe that across semantic-enhanced tokenizers, LM-SPT achieves substantially higher UNMI, suggesting more consistent preservation of utterance-level linguistic content.

\bgroup
\def\arraystretch{1.2}%
\begin{table}[t]
\resizebox{.48\textwidth}{!}{
    \begin{threeparttable}
    \centering
    \caption{\label{tab:human_eval} Human Evaluation Results.}
    \small

    \begin{tabular}{l c cc cc}
    \toprule
    \multirow{3}{*}{Model} &
    \multicolumn{1}{c}{\emph{Reconstruction}} &
    \multicolumn{4}{c}{\emph{Zero-shot Text-to-Speech}} \\
    \cmidrule(lr){2-2}\cmidrule(lr){3-6}
    & \multirow{2}{*}{MUSHRA $\uparrow$} &
    \multicolumn{2}{c}{English} &
    \multicolumn{2}{c}{Korean} \\
    \cmidrule(lr){3-4}\cmidrule(lr){5-6}
    & &
    NMOS $\uparrow$& SMOS $\uparrow$&
    NMOS $\uparrow$& SMOS $\uparrow$\\
    \midrule
    CosyVoice2~\cite{cosyvoice2} & -- & \hl{2.84 $\pm 0.32$} & \hl{2.95 $\pm 0.18$} & \hl{3.22 $\pm 0.24$} & \hl{2.73 $\pm 0.30$} \\
    Mimi~\cite{moshi} & \hl{74.08 $\pm 5.43$} & \hl{2.96 $\pm 0.28$} & \hl{\underline{3.73} $\pm 0.22$} & \hl{3.10 $\pm 0.34$} & \hl{3.38 $\pm 0.22$} \\
    DualCodec~\cite{dualcodec} & \hl{\underline{86.69} $\pm 3.26$} & \hl{\underline{3.54} $\pm 0.27$} & \hl{3.67 $\pm 0.19$} & \hl{\underline{3.62} $\pm 0.26$} & \hl{\textbf{3.80} $\pm 0.28$} \\
    LM-SPT & \hl{\textbf{89.96} $\pm 3.99$} & \hl{\textbf{3.89} $\pm 0.25$} & \hl{\textbf{3.91} $\pm 0.20$} & \hl{\textbf{3.87} $\pm 0.25$} & \hl{\underline{3.79} $\pm 0.27$} \\
    \bottomrule
    \end{tabular}

    \begin{tablenotes}[flushleft]
        \footnotesize
        \item[1] \textit{The \textbf{bold} text and \underline{underscored} text represent the best and second-best results.}
        \item[2] \textit{\hl{Scores are reported as mean $\pm$ 95\% confidence interval. Confidence intervals are computed over listener-level mean scores.}}
    \end{tablenotes}
    \end{threeparttable}
}
\end{table}
\egroup

\subsection{Speech Language Model Performance}
In ASR tasks, we evaluate the performance using \textit{word error rate (WER)} and \textit{character error rate (CER)} for English and Korean, respectively. 
In TTS tasks, we evaluate the performance using four metrics: \textit{word error rate (WER)} for English and \textit{character error rate (CER)} for Korean, which measures speech intelligibility using \textit{Whisper large-v3}~\cite{whisper} model \hl{as well as Wav2Vec2~\cite{wav2vec2}-based ASR models for English\textsuperscript{\ref{fn:w2v2en}}
 (denoted as \textit{Wav2vec2-En}) and Korean\footnote{\url{https://huggingface.co/kresnik/wav2vec2-large-xlsr-korean}}
 (denoted as \textit{Wav2vec2-Ko})}, \textit{UTMOS}~\cite{saeki2022utmos}, which assesses naturalness and overall audio quality, \textit{DNSMOS}~\cite{DNSMOS}, which evaluates noise suppression effectiveness, and \textit{speaker cosine similarity (SIM)}, which measures how closely the generated speech matches the speaker identity using \textit{ERes2NetV2}~\cite{3d-speaker}.
We use \textit{LibriSpeech}~\cite{librispeech} and \textit{KSponSpeech}~\cite{KsponSpeech} as benchmarks.

The results are shown in ~\tabref{tab:main_slm}.
Overall, LM-SPT achieves the best quantitative performance across both ASR and TTS, consistently outperforming other baselines.
Despite being built upon a pretrained LM-aligned speech encoder, CosyVoice2~\cite{cosyvoice2} performs poorly on ASR compared to other tokenizers, consistent with observations in prior work~\cite{stabletoken}. 
Since it remains \emph{semantic--acoustic entangled}, its discrete representations can vary substantially with acoustic factors even when the underlying semantic content is unchanged, which may make them harder for an LLM to interpret and generalize over. 
We also observe that CosyVoice2 attains a notably higher DNSMOS score on TTS compared to other tokenizers.
However, this improvement may stem from the characteristics of its semantic-focused tokenization, which can discard diverse acoustic information (including noise) during encoding. 
As a result, speaker characteristics are less well preserved, as reflected by lower speaker similarity (SIM).

\subsection{Human Evaluation}
We conduct human evaluations to assess both codec reconstruction quality and TTS performance. For all subjective tests, we randomly sample \hl{15 utterances and collect ratings from 15} expert listeners per test. \hl{We report mean scores with 95\% confidence intervals computed over listener-level mean scores.}
For reconstruction performance, we adopt the MUSHRA protocol~\cite{itur_bs1534_2015} and ask listeners to evaluate the perceptual audio quality of reconstructed samples on a 0–100 scale. In this evaluation, we compare LM-SPT with DualCodec and Mimi.
For TTS, we evaluate synthesized speech using two subjective metrics: Naturalness Mean Opinion Score (NMOS) and Similarity Mean Opinion Score (SMOS), both measured on a 1–5 scale. NMOS reflects how natural and human-like the synthesized speech sounds, while SMOS measures how closely the generated speech resembles the target speaker’s voice. In the TTS evaluation, we compare LM-SPT with CosyVoice2, DualCodec, and Mimi for both English and Korean.

The results are shown in Table~\ref{tab:human_eval}. LM-SPT \hl{shows favorable and competitive} subjective performance across both reconstruction and TTS evaluations. For reconstruction, \hl{LM-SPT obtains a MUSHRA score comparable to DualCodec and higher than Mimi, suggesting that the proposed LM-aligned distillation preserves competitive perceptual reconstruction quality.} For TTS, LM-SPT consistently achieves the \hl{highest mean NMOS} in both English and Korean, demonstrating its effectiveness for language-model-based speech generation. \hl{Regarding speaker similarity, LM-SPT obtains the highest SMOS in English, while its Korean SMOS remains close to DualCodec, which achieves the highest score in that setting. Overall, these results suggest that LM-SPT provides favorable perceived naturalness  while maintaining competitive speaker similarity across languages.}

\bgroup
\def\arraystretch{1.25}%
\begin{table*}[t]
\resizebox{\textwidth}{!}{
    \begin{threeparttable}
    \centering
    \caption{\label{tab:main_abl} Ablation Results. Signal-level and Application-level scores are obtained from Codec-SUPERB~\cite{codec-superb}. UNMI denotes utterance-normalized mutual information (see details in~\subsecref{sec:exp_unmi}). Super-scripted by * as the default setting for LM-SPT.}
    \small

    \begin{tabular}{l*{13}{c}}
    \toprule
    \multirow{4}{*}{Model}
    & \multicolumn{8}{c}{\textit{Codec Performance}}
    & \multicolumn{5}{c}{\textit{SLM Performance}} \\
    \cmidrule(lr){2-9}\cmidrule(lr){10-14}

    & \multicolumn{3}{c}{\textit{Signal-level}}
    & \multicolumn{4}{c}{\textit{Application-level}}
    & \multicolumn{1}{c}{\textit{Semantic-level}}
    & \multicolumn{1}{c}{\textit{ASR}}
    & \multicolumn{4}{c}{\textit{Zero-shot TTS}} \\
    \cmidrule(lr){2-4}\cmidrule(lr){5-8}\cmidrule(lr){9-9}\cmidrule(lr){10-10}\cmidrule(lr){11-14}

    & SDR $\uparrow$ & Mel loss $\downarrow$ & PESQ $\uparrow$
    & ACC $\uparrow$ & EER $\downarrow$ & WER $\downarrow$ & ACC $\uparrow$
    & \multirow{2}{*}{UNMI $\uparrow$}
    & \multirow{2}{*}{WER $\downarrow$}
    & \multirow{2}{*}{WER $\downarrow$} & \multirow{2}{*}{UTMOS $\uparrow$} & \multirow{2}{*}{DNSMOS $\uparrow$} & \multirow{2}{*}{SIM $\uparrow$} \\
    &  &  &
    & (ER) & (ASV) & (ASR) & (EC)
    &  &  &  &  &  &  \\
    \midrule
    \multicolumn{14}{l}{\textbf{\textit{Semantic distillation strategy}}}\\
    Feature-level & 2.54 & 0.85 & 1.93 & \textbf{73.89} & 3.71 & 4.88 & 49.80 & 0.7063 $\pm{0.039}$ & 28.07 & 11.97 & 3.76 & 3.31 & 0.69  \\
    Feature-level reconstruction & 2.91 & 1.01 & 1.91 & 72.22 & 4.95 & 4.19 & 46.25 & 0.7897 $\pm{0.008}$ & \textbf{13.24} & 4.95 & 3.84 & \textbf{3.36} & 0.72 \\
    Semantic speech-resynthesis* & \textbf{5.53} & \textbf{0.75} & \textbf{2.20} & 73.61 & \textbf{2.84} & \textbf{3.73} & \textbf{57.00} & \textbf{0.7938} $\pm 0.004$ & 15.50 & \textbf{4.04} & \textbf{4.17} & \textbf{3.36} & \textbf{0.75} \\
    \hline
    \multicolumn{14}{l}{\textbf{\textit{Auxiliary semantic decoder type}}}\\
    Shared decoder & 5.58 &	0.80 & 2.13 & 72.24 & 3.81 & 3.49 & 49.79 & 0.7295 $\pm 0.028$ & 18.08 & 8.90 & 3.77 & 3.19 & 0.64  \\
    w/o gradient update & \textbf{6.83} & \textbf{0.71} & \textbf{2.31} & \textbf{73.82} & \textbf{2.68} & \textbf{3.19} &	\textbf{57.41} & 0.6888 $\pm 0.023$ & 98.28 & 39.23 & 3.34 & 2.91 & 0.51 \\
    Decoupled decoder* & 5.53 & 0.75 & 2.20 & 73.61 & 2.84 & 3.73 & 57.00 & \textbf{0.7938} $\pm 0.004$ & \textbf{15.50} & \textbf{4.04} & \textbf{4.17} & \textbf{3.36} & \textbf{0.75} \\
    \hline
    \multicolumn{14}{l}{\textbf{\textit{Auxiliary semantic decoder capacity}}} \\
    40.5M & 3.84 & 0.86 & 1.99 & 70.9 & 4.33 & 4.28 & 50.45 & 0.7932 $\pm 0.004$ & 21.78 & 4.98 & 4.16 & 3.35 & 0.74 \\
    1.2M* & \textbf{5.53} & \textbf{0.75} & \textbf{2.20} & \textbf{73.61} & \textbf{2.84} & \textbf{3.73} & \textbf{57.00} & \textbf{0.7938} $\pm 0.004$ & \textbf{15.50} & \textbf{4.04} & \textbf{4.17} & \textbf{3.36} & \textbf{0.75} \\
    \hline
    \multicolumn{14}{l}{\textbf{\textit{Encoder type}}} \\
    Shared encoder & -3.81 & 0.98 & 1.66 & 70.21 & 7.00 & 4.28 & 40.75 & 0.7877 $\pm 0.005$ & 15.84 & 5.11 & 3.66 & 3.33 & 0.67 \\
    Dual encoder* & \textbf{5.53} & \textbf{0.75} & \textbf{2.20} & \textbf{73.61} & \textbf{2.84} & \textbf{3.73} & \textbf{57.00} & \textbf{0.7938} $\pm 0.004$ & \textbf{15.50} & \textbf{4.04} & \textbf{4.17} & \textbf{3.36} & \textbf{0.75} \\
    \bottomrule
    \end{tabular}

    \begin{tablenotes}
        \item[1] \textit{The \textbf{bold} text and \underline{underscored} text represent the best and second-best results.}
    \end{tablenotes}
    \end{threeparttable}
}
\end{table*}
\egroup


\section{Ablation Study} \label{sec:study}
In this section, we conduct a series of ablation studies to validate the core design choices of LM-SPT. 
All ablation experiments are performed at a 12.5Hz frame rate and use a 0.5B-scale LLM (\textit{Qwen2.5-0.5B-Instruct}~\cite{qwen25}) for downstream evaluation.
Due to the limited computational resources, we use 1K-hour LibriSpeech~\cite{librispeech} dataset for both codec and SLM training in this section.
The codecs are trained for 20 epochs and the learning rate is scheduled with cosine decay from $2\times10^{-4}$ to a minimum of $0$, using 8 NVIDIA A100 GPUs.
For TTS, SLMs are trained for approximately 5 epochs with a constant learning rate of $2\times10^{-4}$, using 4 NVIDIA A100 GPUs.
For ASR, SLMs are trained for approximately 10 epochs with a constant learning rate of $5\times10^{-4}$, using 4 NVIDIA A100 GPUs. 
We keep the rest of the setup unchanged from \secref{sec:experiments} and use \textit{LibriSpeech}~\cite{librispeech} \textit{test-clean} set as benchmark.

\subsection{Effect of Distillation Strategy} \label{sec:abl_strategy}
While the pretrained LM-aligned teacher is primarily used in our \emph{semantic speech-resynthesis} distillation (\figref{fig:schematic_c}), it can also be applied to both the \emph{feature-level} distillation (\figref{fig:schematic_a}) and \emph{feature-level reconstruction} distillation (\figref{fig:schematic_b}).
Here, we empirically compare our proposed distillation strategy against the feature-level variant under the same LM-aligned teacher (i.e., \emph{Whisper-large-v3}).
Specifically, for \emph{feature-level distillation}, we replace the original SSL teacher with the Whisper-large-v3 encoder and minimize the cosine distance between teacher and student representations, following Mimi.
For \emph{feature-level reconstruction distillation}, we likewise replace the pretrained SSL-based semantic encoder with the Whisper-large-v3 encoder and optimize the semantic feature reconstruction objective using an MSE loss, following DualCodec.
For a fair comparison, we keep the encoder/decoder configurations, codebook sizes, and other codec hyperparameters identical to those of LM-SPT.

As shown in the 1st upper part of \tabref{tab:main_abl}, we observe that our \emph{semantic speech-resynthesis} distillation consistently outperforms both feature-level distillation and feature-level reconstruction distillation on the codec benchmark as well as on TTS, indicating that resynthesis-based supervision provides a stronger learning signal for semantic distillation.
As expected, the \emph{feature-level reconstruction} strategy exhibits particularly strong performance on the ASR downstream task, since it directly exploits a pretrained ASR encoder (i.e., Whisper-large-v3) as the dedicated semantic pathway.
However, this gain comes with a practical constraint: it requires a relatively large semantic encoder, as it relies on a pretrained teacher encoder for audio tokenization at test time (in our setting, we use the \emph{Whisper-large-v3} encoder, resulting in a 664.6M-parameter semantic encoder, whereas LM-SPT's semantic encoder has 16.2M parameters), which can be prohibitive for lightweight or on-device SLM deployments.

\subsection{Impact of Auxiliary Semantic Decoder} \label{sec:abl_decoder}
As discussed in \subsecref{method:decoup}, using a shared decoder for both reconstruction and semantic distillation can lead to interference between objectives, potentially degrading both the reconstruction fidelity and semantic representation power.
We empirically observe this interference in the results denoted as \textit{auxiliary semantic decoder type} in 2nd upper part of \tabref{tab:main_abl}.
A possible solution is to freeze the decoder parameters during semantic distillation (i.e., without gradient updates from the distillation loss).
While this alleviates degradation in reconstruction fidelity, it significantly weakens the ability to learn meaningful semantic representations.
This setting is denoted as \textit{w/o gradient update} in the results.
In contrast, our proposed design, which uses a decoupled semantic decoder, effectively mitigates both issues, providing a better balance between the two objectives.

Furthermore, as also discussed in \subsecref{method:decoup}, the capacity of the decoupled decoder plays a key role in acting as an information bottleneck.
As shown in 3rd upper part (denoted as \textit{auxiliary semantic decoder capacity}) of \tabref{tab:main_abl}, using a lightweight decoder (1.2M) substantially improves both codec and downstream performances, suggesting that a tighter bottleneck encourages the encoder to learn more effective semantic and acoustic representations.

\subsection{Impact of Dual Encoder} \label{sec:abl_encoder}
As discussed in \subsecref{method:dual_enc}, we adopt a dual encoder architecture to capture high-level semantic information and low-level acoustic details without conflict.
To verify the effect of this design, we additionally train LM-SPT with a single \emph{shared} encoder (i.e., using one encoder to produce both semantic and acoustic latents) while keeping the rest of the setup unchanged.
The results are shown in the lower part (denoted as \textit{encoder type}) of \tabref{tab:main_abl}.
We observe that the dual-encoder variant outperforms the shared-encoder counterpart not only on codec performance but also on downstream SLM evaluations.
These results suggest that explicitly separating the encoding pathways helps mitigate the conflicting objectives in semantic-enhanced tokenizers, i.e., semantic distillation and waveform reconstruction.

\section{Theoretical Justification} \label{sec:theory}
\hl{
To clarify why semantic speech-resynthesis distillation can be advantageous over conventional feature-level distillation, we present a theoretical comparison under simplified probabilistic assumptions.
}
\hl{
\subsection{Setup} \label{sec:theory_setup}
We frame both feature-level and semantic speech-resynthesis distillation as KL minimization between multivariate Gaussian distributions over the encoder output $f_T(x), f_S(x) \in \mathbb{R}^d$:
\begin{align}
\label{eq:loss_to_kl_combined}
\begin{split}
    \mathcal{L}_\text{distill}^\text{feat} &= \operatorname{MSE}\big(f_S(x), f_T(x)\big) \;\propto\; \operatorname{D}_{\mathrm{KL}}\big(P_{f_T}(x) \,\|\, P_{f_S}(x)\big), \\
    \mathcal{L}_\text{distill}^\text{resyn} &= \operatorname{MSE}\big(f_T(\hat{x}), f_T(x)\big) \;\propto\; \operatorname{D}_{\mathrm{KL}}\big(P_{f_T}(x) \,\|\, P_{f_T}(\hat{x})\big),
\end{split}
\end{align}
where $f_S$ denotes the student semantic pathway trained to approximate $f_T$, and $\hat{x}$ denotes the reconstructed output of $x$. 
Here, $\hat{x}$ is used as a generic notation for a reconstructed waveform. In LM-SPT, this reconstruction is instantiated as the semantic-only resynthesized waveform $\hat{x}_\text{sem}=\mathcal{G}_\text{aux}(z_\text{sem})$ in~\equref{eq:distill}, while the KL-based argument applies more generally to any reconstruction whose teacher-encoder features preserve the underlying semantic content.

We model the distributions as
\begin{align*}
\begin{split}
    P_{f_T}(x) &= \mathcal{N}(\mu_{f_T}, \Sigma_{f_T}), \\
    P_{f_S}(x) &= \mathcal{N}(\mu_{f_S}, \Sigma_{f_S}), \\
    P_{f_T}(\hat{x}) &= \mathcal{N}(\hat{\mu}_{f_T}, \hat{\Sigma}_{f_T}),
\end{split}
\end{align*}
with positive-definite covariances $\Sigma_{f_T}, \Sigma_{f_S}, \hat{\Sigma}_{f_T} \in \mathbb{R}^{d \times d}$.
Each KL divergence in~\equref{eq:loss_to_kl_combined} decomposes into a mean-dependent term and a covariance-dependent term: 
\begin{equation}
\begin{aligned}
\label{eq:kl_decompose}
\operatorname{D}_{\mathrm{KL}}\!\big(& P_{f_T(x)} \,\Vert\, \mathcal{N}(\mu, \Sigma)\big) \\
& = \underbrace{\tfrac{1}{2}(\mu - \mu_{f_T})^{\!\top}\!\Sigma^{-1}(\mu - \mu_{f_T})}_{\text{mean-dependent term}} \\
& + \underbrace{\tfrac{1}{2}\!\left[\operatorname{tr}(\Sigma^{-1}\Sigma_{f_T}) - d + \log\det(\Sigma\,\Sigma_{f_T}^{-1})\right]}_{\text{covariance-dependent term}}.
\end{aligned}
\end{equation}
}
\hl{
\subsection{Main Result} \label{sec:theory_main}
In an idealized setting in which the training data and capacity are sufficient to reach near-optimal solutions, the following approximations are expected to hold at convergence,
\begin{align}
\begin{split}
    \mu_{f_S} &= \mathbb{E}_x\big[f_S(x)\big] \approx \mathbb{E}_x\big[f_T(x)\big], \\
    \hat{\mu}_{f_T} &= \mathbb{E}_{\hat{x}}\big[f_T(\hat{x})\big] \approx \mathbb{E}_x\big[f_T(x)\big],
\end{split}
\end{align}
so that the mean-dependent term in~\equref{eq:kl_decompose} vanishes for both cases of~\equref{eq:loss_to_kl_combined}, and the divergence reduces to the covariance-dependent term, which admits an equivalent eigenvalue form:
\begin{align}
\label{eq:kl_cov_def}
\begin{split}
    \mathrm{KL}_{\mathrm{cov}}\big(\Sigma_{f_T}, \Sigma\big)
    &= \tfrac{1}{2}\Big[
        \operatorname{tr}\!\big(\Sigma^{-1}\Sigma_{f_T}\big) - d + \log\det\!\big(\Sigma\,\Sigma_{f_T}^{-1}\big)
    \Big] \\
    &= \tfrac{1}{2}\sum_{i=1}^{d}\big(\lambda_i - 1 - \log \lambda_i\big),
\end{split}
\end{align}
where $\{\lambda_i\}_{i=1}^{d}$ are the eigenvalues of $\Sigma^{-1}\Sigma_{f_T}$. Since $g(\lambda) = \lambda - 1 - \log \lambda \ge 0$ for all $\lambda > 0$ with equality only at $\lambda = 1$, $\mathrm{KL}_{\mathrm{cov}}$ is non-negative and vanishes if and only if $\Sigma = \Sigma_{f_T}$.
}

\hl{
\begin{lemma}[Variance shrinkage of the MSE-trained student]\label{lem:var_shrink}
Let $Z = z_\text{sem}(x)$ denote the student bottleneck representation. 
The MSE-optimal student predictor under $Z$ is the conditional mean $f_S(x) = \mathbb{E}\big[f_T(x) \mid Z\big]$, so the student covariance equals the covariance of this conditional mean:
$\Sigma_{f_S} = \operatorname{Cov}\!\big(\mathbb{E}[f_T(x) \mid Z]\big)$.
Applying the law of total covariance to $f_T(x)$ and identifying its second term with $\Sigma_{f_S}$,
\begin{align}
    \Sigma_{f_T} = \mathbb{E}\big[\operatorname{Cov}(f_T(x) \mid Z)\big] + \Sigma_{f_S}
    \;\;\Longrightarrow\;\;
    \Sigma_{f_S} \preceq \Sigma_{f_T}.
\end{align}
This implies that every eigenvalue $\lambda_i$ of $\Sigma_{f_S}^{-1}\Sigma_{f_T}$ satisfies $\lambda_i \ge 1$, with strict inequality in directions where $Z$ fails to fully determine $f_T(x)$. 
Hence $\mathrm{KL}_{\mathrm{cov}}(\Sigma_{f_T}, \Sigma_{f_S}) > 0$ generically.
\end{lemma}
}

\hl{
\begin{remark}[Resynthesis approximately preserves the teacher covariance]\label{remark}
Since $f_T(\hat{x})$ is produced by the same frozen encoder applied to a perceptually similar waveform $\hat{x}$, and LM-aligned encoders are empirically insensitive to waveform perturbations when the underlying semantic content is unchanged (e.g., different realizations of the same utterance), we expect $\hat{\Sigma}_{f_T} \approx \Sigma_{f_T}$ in the full-covariance sense.
Equivalently, the eigenvalues $\tilde{\lambda}_i$ of $\hat{\Sigma}_{f_T}^{-1}\Sigma_{f_T}$ tend to be close $\tilde{\lambda}_i \approx 1$, and thus $\mathrm{KL}_{\mathrm{cov}}(\Sigma_{f_T}, \hat{\Sigma}_{f_T}) \approx 0$.
We empirically validate this assumption on a trained LM-SPT in~\subsecref{sec:remark1_empirical}.
\end{remark}
}

\hl{
Combining Lemma~\ref{lem:var_shrink} and Remark~\ref{remark} yields
\begin{align} \label{eq:kl_gap}
    \operatorname{D}_{\mathrm{KL}}\!\big(P_{f_T}(x) \,\|\, P_{f_S}(x)\big)
    \;\gtrsim\;
    \operatorname{D}_{\mathrm{KL}}\!\big(P_{f_T}(x) \,\|\, P_{f_T}(\hat{x})\big),
\end{align}
with the gap strictly positive whenever feature-level distillation incurs any non-trivial covariance shrinkage. 
Semantic speech-resynthesis distillation can therefore achieve essentially zero KL divergence under reasonable assumptions, offering a stronger form of semantic alignment when paired with an LM-aligned teacher. 
We provide direct empirical support for this prediction in \subsecref{sec:abl_strategy} by comparing other distillation strategies using the same teacher.
}

\hl{
\subsection{Extension to Cosine-Based Distillation}
The analysis above is formulated under MSE, while feature-level distillation pipelines (e.g.,~\cite{zhang2024speechtokenizer,moshi}) typically use a cosine similarity loss.
Since our objective is to compare distillation strategies on a common ground — namely, the KL divergence between teacher and student feature distributions induced by each loss — we evaluate $f_S^{\cos}$ within the same KL framework rather than within $\mathcal{L}_{\cos}$ itself. Under this common-metric view, we show that (i) the shrinkage-induced covariance mismatch of Lemma~\ref{lem:var_shrink} persists under cosine loss, and (ii) cosine loss additionally incurs a magnitude-mismatch penalty.
}

\noindent
\hl{
\textbf{Matched-magnitude regime.} The MSE can be decomposed as follows
\begin{align}
\label{eq:mse_decomp_combined}
    \operatorname{MSE}\big(f_S(x),f_T(x)\big) &= \underbrace{\big(\lVert f_T(x) \rVert - \lVert f_S(x) \rVert\big)^2}_{\text{magnitude-mismatch term}} \nonumber\\
    & + \underbrace{2\,\lVert f_T(x) \rVert \lVert f_S(x) \rVert\,(1 - \cos\theta)}_{\text{direction-mismatch term}},
\end{align}
where $\cos\theta = \frac{f_T(x)^\top f_S(x)}{\|f_T(x)\|\|f_S(x)\|}$.
The cosine loss is defined as $\mathcal{L}_{\cos} \coloneqq 1 - \cos\theta$, which penalizes only the direction mismatch, whereas MSE penalizes both magnitude and direction mismatches.
When $\|f_T(x)\| \approx \|f_S(x)\| = c$, the magnitude term in~\equref{eq:mse_decomp_combined} vanishes and $\operatorname{MSE} \approx 2c^{2}\mathcal{L}_{\cos}$. Together with the MSE--KL proportionality of~\equref{eq:loss_to_kl_combined}, this yields
\begin{equation}
\mathcal{L}_{\cos}(f_T(x), f_S(x))
\;\propto\;
\tfrac{1}{2c^{2}}\, \operatorname{D}_{\mathrm{KL}}\!\big(P_{f_T(x)} \,\|\, P_{f_S(x)}\big),
\label{eq:cos_kl}
\end{equation}
thus cosine and MSE share the same minimizer, and the KL behavior described in~\equref{eq:kl_gap} can be applied in the same way.
}

\noindent
\hl{
\textbf{General regime.} $\mathcal{L}_{\cos}$ is invariant to the student's magnitude, so the cosine-optimal student $f_S^{\cos}$ eliminates only the direction-mismatch term in~\equref{eq:mse_decomp_combined}, whereas the MSE-optimal student $f_S$ eliminates both terms jointly.
Hence,
\begin{equation}
\operatorname{MSE}(f_T(x), f_S^{\cos}(x))
\;\geq\;
\operatorname{MSE}(f_T(x), f_S(x)).
\label{eq:cos_residual}
\end{equation}
By the MSE--KL proportionality of~\equref{eq:loss_to_kl_combined}, the cosine-optimal student therefore incurs
$\operatorname{D}_{\mathrm{KL}}(P_{f_T(x)} \,\|\, P_{f_S^{\cos}(x)}) \geq \operatorname{D}_{\mathrm{KL}}(P_{f_T(x)} \,\|\, P_{f_S(x)})$.
}


\noindent
\hl{
This, together with~\equref{eq:kl_gap} extends the KL gap to
\begin{align}
\underbrace{\operatorname{D}_{\mathrm{KL}}\!\big(P_{f_T(x)} \,\|\, P_{f_S^{\cos}(x)}\big)}_{\text{cosine-based distillation}}
&\;\geq\;
\underbrace{\operatorname{D}_{\mathrm{KL}}\!\big(P_{f_T(x)} \,\|\, P_{f_S(x)}\big)}_{\text{MSE-based distillation}} \notag \\
&\;\gtrsim\;
\underbrace{\operatorname{D}_{\mathrm{KL}}\!\big(P_{f_T(x)} \,\|\, P_{f_T(\hat{x})}\big)}_{\text{resynthesis-based distillation (ours)}},
\label{eq:kl_gap_full}
\end{align}
where the first inequality is strict whenever magnitudes are mismatched, and the second is strict whenever feature-level distillation incurs the covariance shrinkage of Lemma~\ref{lem:var_shrink}.
}

\begin{figure*}[t]
    \centering
    \subfloat[ch.\,82]{%
        \includegraphics[width=0.24\linewidth]{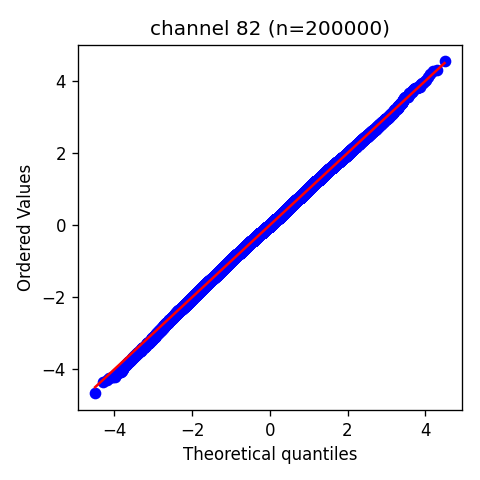}
        \label{fig:qq_ch0082}
    }\hfill
    \subfloat[ch.\,286]{%
        \includegraphics[width=0.24\linewidth]{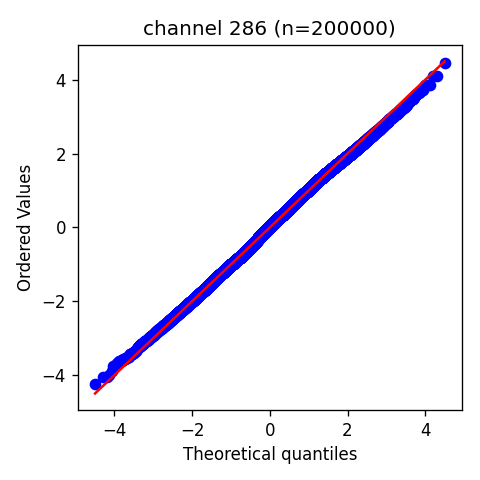}
        \label{fig:qq_ch0286}
    }\hfill
    \subfloat[ch.\,530]{%
        \includegraphics[width=0.24\linewidth]{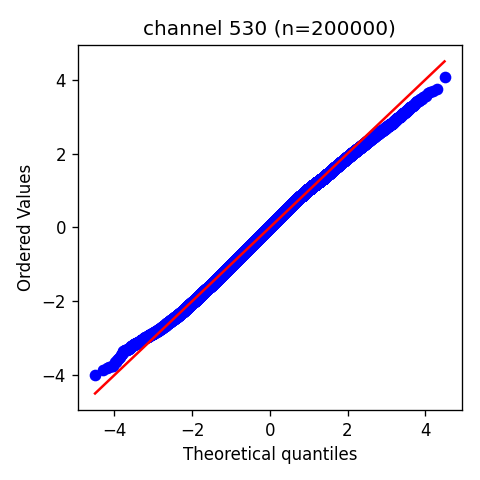}
        \label{fig:qq_ch0530}
    }\hfill
    \subfloat[ch.\,621]{%
        \includegraphics[width=0.24\linewidth]{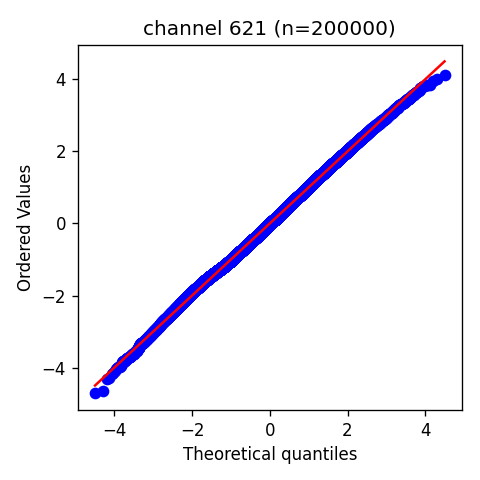}
        \label{fig:qq_ch0621}
    }\hfill \\
    \subfloat[ch.\,788]{%
        \includegraphics[width=0.24\linewidth]{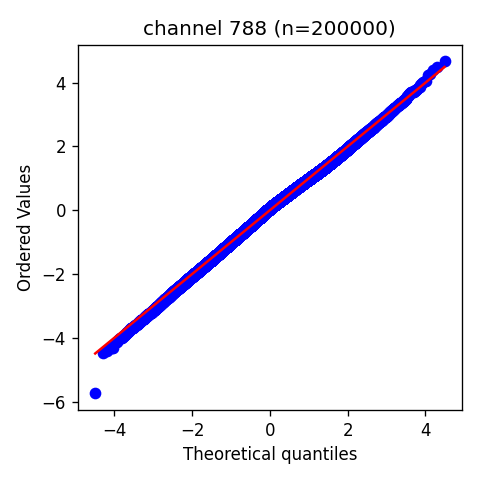}
        \label{fig:qq_ch0788}
    }\hfill
    \subfloat[ch.\,861]{%
        \includegraphics[width=0.24\linewidth]{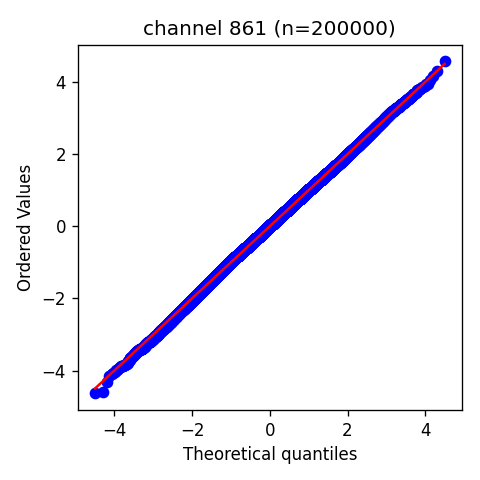}
        \label{fig:qq_ch0861}
    }\hfill
    \subfloat[ch.\,995]{%
        \includegraphics[width=0.24\linewidth]{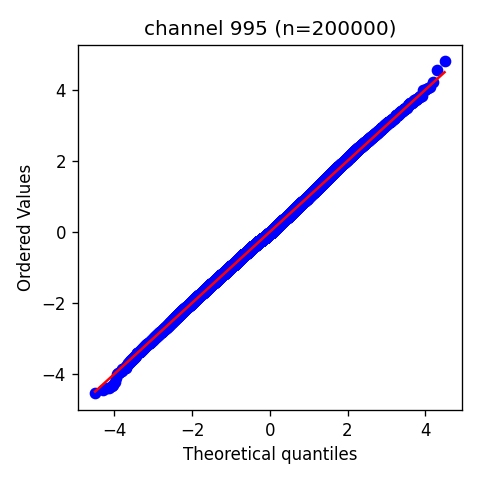}
        \label{fig:qq_ch0995}
    }\hfill
    \subfloat[ch.\,1047]{%
        \includegraphics[width=0.24\linewidth]{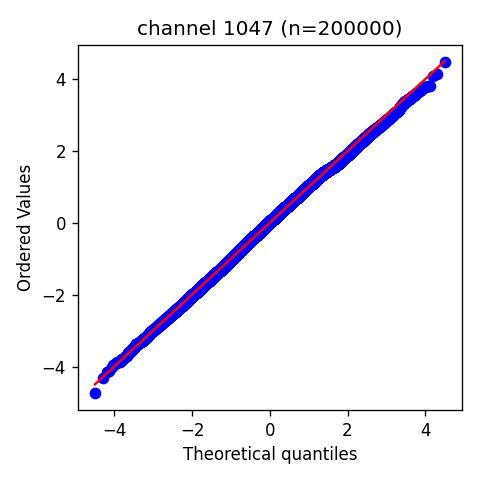}
        \label{fig:qq_ch1047}
    }\hfill
    \vspace{0.5em}
    \caption{
        Per-channel QQ plots of Whisper-large-v3 encoder features $f_T(x)$ against the standard normal $\mathcal{N}(0,1)$. Eight channels (a)--(h) are \emph{randomly sampled} from the $d{=}1280$ channels of the encoder. Each subplot plots the empirical quantiles (blue dots) against theoretical Gaussian quantiles (red line) after per-channel standardization. Across the sampled channels, the empirical quantiles fall close to the diagonal over $\pm 4\sigma$, providing direct visual support for the marginal Gaussian approximation used in our analysis.
    }
    \label{fig:whisper_qq}
\end{figure*}

\begin{figure}[t]
    \centering
    \subfloat[Eigenvalues of $\hat{\Sigma}_{f_T}^{-1}\Sigma_{f_T}$]{%
        \includegraphics[width=0.96\linewidth]{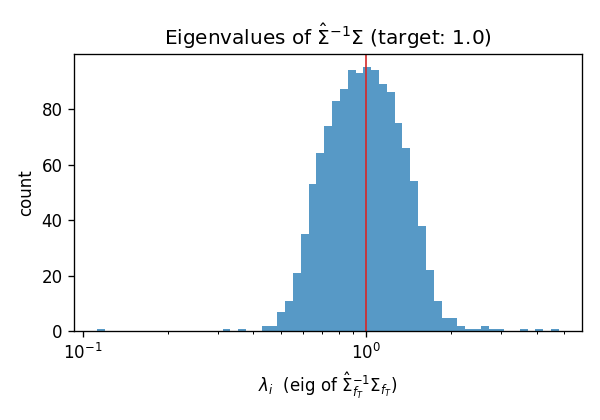}
        \label{fig:remark1_lambda}
    }
    \\
    \subfloat[Eigenvalue spectra of $\Sigma_{f_T}$ vs $\hat{\Sigma}_{f_T}$]{%
        \includegraphics[width=0.96\linewidth]{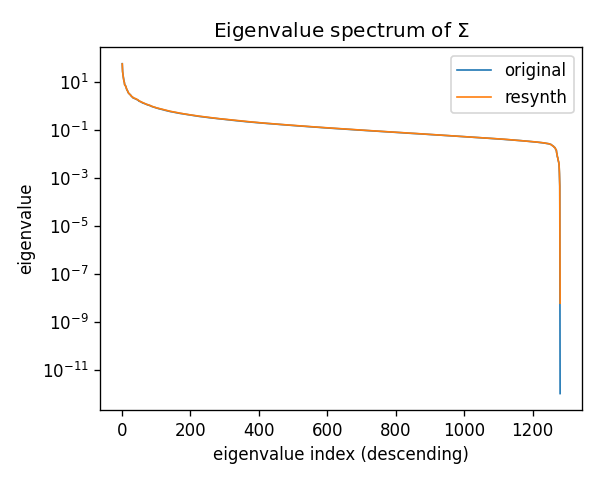}
        \label{fig:remark1_spectrum}
    }
    \caption{
        Empirical validation of Remark~\ref{remark} on a trained LM-SPT model.
        (a) The eigenvalues of $\hat{\Sigma}_{f_T}^{-1}\Sigma_{f_T}$ form a tight unimodal distribution around $1$ (median $0.985$; $98\%$ of directions in $[0.5, 2]$).
        (b) The eigenvalue spectra of the original-audio and semantic-resynthesized-audio covariances overlap across the full $d{=}1280$ range.
    }
    \label{fig:remark1_validation}
\end{figure}

\hl{
\subsection{On the Gaussian Approximation}
Our KL-based comparison relies on a Gaussian approximation to obtain a closed-form decomposition into mean and covariance terms and to make the resulting gap explicit and interpretable.
To justify that this approximation is reasonable in our setting, we provide an empirical sanity check verifying multivariate Gaussianity on the feature distribution induced by our training pipeline.
On $5{,}715$ utterances from the GigaSpeech~\cite{GigaSpeech} validation set, we randomly crop a $6$-second segment (matching the codec training crop length) for each utterance, pass it through Whisper-large-v3, and retain only encoder frames covering the active region (discarding the silence frames introduced by Whisper's internal zero-padding to $30$ seconds).
We collect approximately $2\times 10^5$ samples per channel for all $d{=}1280$ channels.
The marginals are strongly Gaussian-like: median $|\text{skewness}|=0.046$ (with 97.3\% of channels having |skewness| < 0.25), median $|\text{excess kurtosis}|=0.048$ (with 96.9\% having |excess kurtosis| < 0.5), and Quantile-Quantile (QQ) plots closely match the standard normal for the vast majority of channels as shown in~\figref{fig:whisper_qq}.
}

\hl{
\subsection{Empirical Validation of Remark 1} \label{sec:remark1_empirical}
On the same setup as above (GigaSpeech validation utterances after $6$-second active-speech crops), we additionally test the assumption that underlies Remark~\ref{remark}. 
For each utterance, we compute Whisper-large-v3 features twice: from the original waveform~$x$ and from a waveform~$\hat{x}$ obtained by re-encoding $x$ through LM-SPT and decoding from the \emph{semantic codebook only} via the auxiliary decoder $\mathcal{G}_\text{aux}$--the path that the distillation loss directly optimizes. 
We use the two populations to estimate the empirical covariances $\Sigma_{f_T}$ and $\hat{\Sigma}_{f_T}$.
Here, as shown in~\figref{fig:remark1_validation}, the two covariances are nearly identical.
The eigenvalues $\tilde{\lambda}_i$ of $\hat{\Sigma}_{f_T}^{-1}\Sigma_{f_T}$ form a tight, unimodal distribution around the target value~$1$ (median $0.985$, mean $1.04$; $98\%$ of directions lie in $[0.5, 2]$).
Per-channel means and standard deviations of $f_T(x)$ and $f_T(\hat{x})$ are almost perfectly aligned (Pearson $r{=}0.998$ and $r{=}1.000$, respectively), and the corresponding eigenvalue spectra are visually indistinguishable.
These observations directly support $\hat{\Sigma}_{f_T} \approx \Sigma_{f_T}$ in Remark~\ref{remark}.
}

\section{Analysis} \label{sec:analysis}

\begin{figure*}[t]
    \centering
    \subfloat[English utterance]{%
        \includegraphics[width=0.39\linewidth]{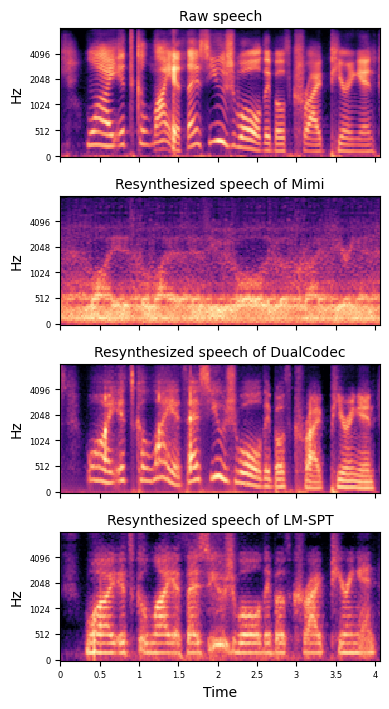}
        \label{fig:mel_eng}
    }
    \hspace{0.04\linewidth}%
    \subfloat[Korean utterance]{%
        \includegraphics[width=0.39\linewidth]{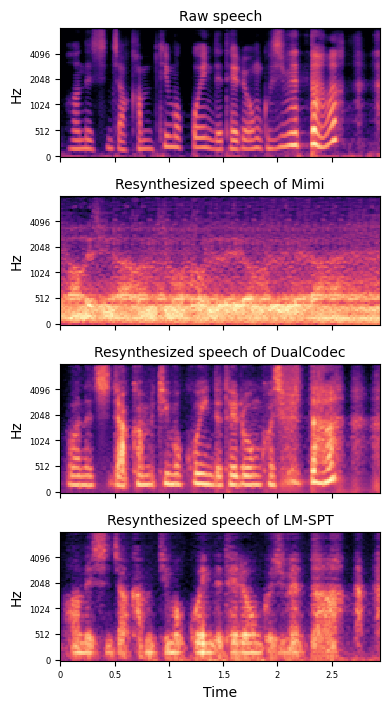}
        \label{fig:mel_kor}
    }
    \caption{Mel-spectrograms of raw speech and reconstructed outputs from three models (LM-SPT, Mimi, DualCodec), using only semantic tokens. DualCodec visually resembles the original signal but likely retains non-semantic cues, while LM-SPT emphasizes semantic fidelity with reduced prosodic and timbral components.}
    \label{fig:mel}
\end{figure*}

\subsection{Semantic--Acoustic Disentanglement}
\bgroup
\def\arraystretch{1.2}%
\begin{table}[h]
\resizebox{0.48\textwidth}{!}{
    \begin{threeparttable}
    \centering
    \caption{\label{tab:semantic_recon} Semantic tokens only reconstruction results.}
    \small

    \begin{tabular}{lcccccc}
    \toprule
    & \multicolumn{3}{c}{LibriSpeech (test-clean)} & \multicolumn{3}{c}{KSponSpeech (test-clean)} \\
    \cmidrule(lr){2-4}\cmidrule(lr){5-7}
    Model 
    & \multicolumn{2}{c}{WER $\downarrow$} & SIM $\downarrow$
    & \multicolumn{2}{c}{CER $\downarrow$} & SIM $\downarrow$ \\
    \cmidrule(lr){2-3}\cmidrule(lr){5-6}
    & \scriptsize \textit{Whisper} & \scriptsize \hl{\textit{Wav2vec2-En}} & 
    & \scriptsize \textit{Whisper} & \scriptsize \hl{\textit{Wav2vec2-Ko}} & \\
    \midrule

    \textit{Ground-truth} & 2.74 & \hl{2.48} & 1.00 & 15.14 & \hl{51.87} & 1.00 \\
    \hdashline

    SpeechTokenizer~\cite{zhang2024speechtokenizer} & 5.85 & \hl{5.80} & 0.40 & 82.03 & \hl{75.10} & 0.22 \\
    Mimi~\cite{moshi} & 45.49 & \hl{78.18} & \textbf{0.19} & 92.85 & \hl{87.48} & \textbf{0.07} \\
    DualCodec~\cite{dualcodec} & 6.46 & \hl{5.99} & 0.61 & 49.52 & \hl{66.72} & 0.43 \\
    LM-SPT & \textbf{4.27} & \hl{\textbf{4.16}} & 0.34 & \textbf{38.86} & \hl{\textbf{62.68}} & 0.19 \\

    \bottomrule
    \end{tabular}

    \end{threeparttable}
}
\end{table}
\egroup

In studies of semantic-acoustic disentangled tokenizers, the expressive power of semantic representations is often analyzed not only through downstream task performance, but also through \emph{semantic-only} waveform reconstruction~\cite{zhang2024speechtokenizer,dualcodec}.
Specifically, by reconstructing speech solely from semantic tokens and evaluating the resulting audio, one can assess whether the representation preserves linguistic content while discarding acoustic factors: better content recoverability with weaker acoustic/speaker cues indicates stronger semantic preservation.

LM-SPT is also trained to provide semantic representations centered on linguistic content via \emph{semantic speech-resynthesis distillation}.
Intuitively, this strategy works because the pretrained LM-aligned speech encoder $f_T$ produces representations that are oriented toward semantic content, rather than fine-grained acoustic details. 
By forcing the resynthesis $\hat{x}_{\text{sem}}$ to depend only on the semantic codebook embedding $z_{\text{sem}}$, we impose a strong bottleneck: $z_{\text{sem}}$ must retain the information that is necessary to reproduce $f_T(x)$, while there is limited capacity to also encode speaker timbre or other paralinguistic factors. 
Therefore, optimizing $\mathcal{L}_\text{distill}^\text{\hl{resyn}}$ encourages $z_{\text{sem}}$ to retain the information needed to reproduce the teacher’s content-centric representation. 

To validate this behavior, we reconstruct speech using only semantic tokens and measure (i) content preservation using content error rates (WER for English and CER for Korean), and (ii) the degree of acoustic/speaker leakage using speaker similarity (SIM) between the ground-truth waveform and the semantic-only reconstructed waveform.
In LM-SPT, we use the main decoder (not the auxiliary decoder $\mathcal{G}_\text{aux}$) for the reconstruction.
As shown in \tabref{tab:semantic_recon}, LM-SPT significantly outperforms other semantic-acoustic disentangled tokenizers in preserving semantic content.
While Mimi~\cite{moshi} yields the lowest speaker similarity, its content preservation is substantially worse than other tokenizers.
Overall, LM-SPT achieves a favorable trade-off: it preserves semantic content strongly while keeping acoustic leakage low, and at the same time attains reconstruction performance comparable to DualCodec as shown in ~\tabref{tab:main_recon}, indicating a more effective separation between semantic and acoustic information.
We also provide qualitative analyses in the following~\subsecref{anal:mel_spec}.

\subsection{Melspectrogram Analysis} \label{anal:mel_spec}
\figref{fig:mel} shows mel-spectrograms of raw speech and semantic-only reconstructions generated by Mimi~\cite{moshi}, DualCodec~\cite{dualcodec}, and LM-SPT. 
The spectrogram from Mimi appears generally blurred, with substantial loss of structure compared to the original. DualCodec produces a spectrogram that closely resembles the raw speech in terms of overall energy distribution and structural patterns. However, since it is also reconstructed solely from semantic tokens, this similarity suggests that the model retains non-semantic information such as timbre and prosody.
In contrast, LM-SPT yields a simpler spectrogram with less prominent horizontal patterns, indicating reduced presence of paralinguistic features. This reflects a more semantic-only reconstruction, highlighting a fundamental distinction from the other models.

\subsection{Inference Speed} \label{anal:speed}
\bgroup
\def\arraystretch{1.2}%
\begin{table}[h]
\resizebox{0.48\textwidth}{!}{
    \centering
    \small
    \begin{threeparttable}
    \caption{Model size and real-time factor (RTF) for encode/decode.}
    \label{tab:rtf_params}
        \begin{tabular}{lcccc}
        \toprule
        Model & \multicolumn{2}{c}{Encode} & \multicolumn{2}{c}{Decode} \\
        \cmidrule(lr){2-3}\cmidrule(lr){4-5}
        & \#Params & RTF $\downarrow$ & \#Params & RTF $\downarrow$ \\
        \midrule
        SpeechTokenizer~\cite{zhang2024speechtokenizer} & 68M  & 0.009 & 35M & 0.004 \\
        Mimi~\cite{moshi}           & 38M  & 0.012 & 40M & 0.004 \\
        DualCodec~\cite{dualcodec}      & 622M & 0.023 & 53M & 0.006 \\
        LM-SPT         & 32M  & 0.014 & 40M & 0.005 \\
        \bottomrule
        \end{tabular}
    \end{threeparttable}
}
\arrayrulecolor{black}
\end{table}
\egroup
\hl{
We evaluate the inference cost of LM-SPT in terms of inference speed, measured by \textit{real-time factor} (RTF), and parameter counts for encoding and decoding across semantic-enhanced speech tokenizers.
RTF measures the ratio of processing time to audio duration. 
As in ~\cite{dualcodec}, we report the amortized RTF over LibriSpeech (test-clean) measured on a single A100 GPU with a batch size of 1.
As shown in \tabref{tab:rtf_params}, LM-SPT is substantially more efficient than DualCodec in both parameter count and inference speed.
In particular, LM-SPT does not rely on a large pretrained semantic encoder, unlike DualCodec, resulting in significantly fewer encoding parameters and a lower encoding RTF.
Compared to Mimi, LM-SPT achieves comparable parameter counts and RTFs for both encoding and decoding, indicating similar runtime efficiency while retaining the benefits of LM-aligned tokenization.
}

\section{Conclusion} \label{conclusion}
In this work, we introduce LM-SPT, a speech tokenization method that enhances semantic alignment with LMs.
Motivated by the limitations of feature-level semantic distillation, we propose a semantic speech-resynthesis distillation strategy that leverages supervision from pretrained LM-aligned teacher embeddings, enabling the model to learn semantically meaningful tokens without relying on rigid temporal alignment.
We empirically demonstrate that LM-SPT consistently outperforms baseline speech tokenizers on ASR and TTS within SLM settings, while maintaining strong reconstruction fidelity.
These results highlight LM-SPT as an effective solution for integrating speech into large-scale LMs.
While LM-SPT shows promising results, we focus on basic speech understanding and generation tasks within SLM settings, leaving extensions to broader instruction-following capabilities (e.g., speech editing) as future work.
Moreover, further improvements are needed to enhance performance at very low frame rates (e.g., 6.25Hz), where the trade-off between compression and LM-alignment without compromising reconstruction fidelity is particularly challenging.

\section*{Acknowledgments}
This work was supported by the National Research Foundation of Korea(NRF) grant funded by the Korea government(MSIT) (No. RS-2024-00353007, 20\%), the Institute of Information \& Communications Technology Planning \& Evaluation(IITP) grant funded by the Korea government(MSIT) (IITP-2026-RS-2024-00436857, ITRC(Information Technology Research Center), 20\%), (No. RS-2019-II190079, Artificial Intelligence Graduate School Program(Korea University), 20\%), (IITP-2026-RS-2025-02304828, artificial intelligence star fellowship support program to nurture the best talents, 20\%), Culture, Sports and Tourism R\&D Program through the Korea Creative Content Agency grant funded by the Ministry of Culture, Sports and Tourism in 2026 (Project Name: Development of Core Technologies for Copyright Verification of AI-Generated and Deepfake Music, Project Number: RS-2025-02216483, Contribution Rate: 20\%), and Kakao.
The authors gratefully acknowledge the Multi-modal Model Training Team at Kakao for their continued support throughout this work.



\bibliographystyle{IEEEtran}
\bibliography{bibi.bib}

\newpage

 




\vfill

\end{document}